\documentclass[10pt,twocolumn,letterpaper]{article}

\usepackage{cvpr}
\usepackage{times}
\usepackage{epsfig}
\usepackage{graphicx}
\usepackage{amsmath}
\usepackage{amssymb}
\usepackage{algorithm,algorithmic}

 \cvprfinalcopy 

\floatstyle{plain}
\newfloat{myalgo}{tbhp}{mya}
\newenvironment{Algorithm}[2][tbh]%
{\begin{myalgo}[#1]
\centering
\begin{minipage}{#2}
\begin{algorithm}[H]}%
{\end{algorithm}
\end{minipage}
\end{myalgo}}

\title{Dense Scattering Layer Removal}

\author{Qiong Yan, Li Xu, Jiaya Jia\\
The Chinese University of Hong Kong\\
{\tt\small {\{qyan,xuli,leojia\}@cse.cuhk.edu.hk}} }

\begin{document}

\maketitle

\begin{abstract}
We propose a new model, together with advanced optimization, to separate a thick scattering media layer from a single natural image. It is able to handle challenging underwater scenes and images taken in fog and sandstorm, both of which are with significantly reduced visibility. Our method addresses the critical issue -- this is, originally unnoticeable impurities will be greatly magnified after removing the scattering media layer -- with transmission-aware optimization. We introduce non-local structure-aware regularization to properly constrain transmission estimation without introducing the halo artifacts. A selective-neighbor criterion is presented to convert the unconventional constrained optimization problem to an unconstrained one where the latter can be efficiently solved.
\end{abstract}

\begin{figure}[t]
\centering
\includegraphics[width=\linewidth]{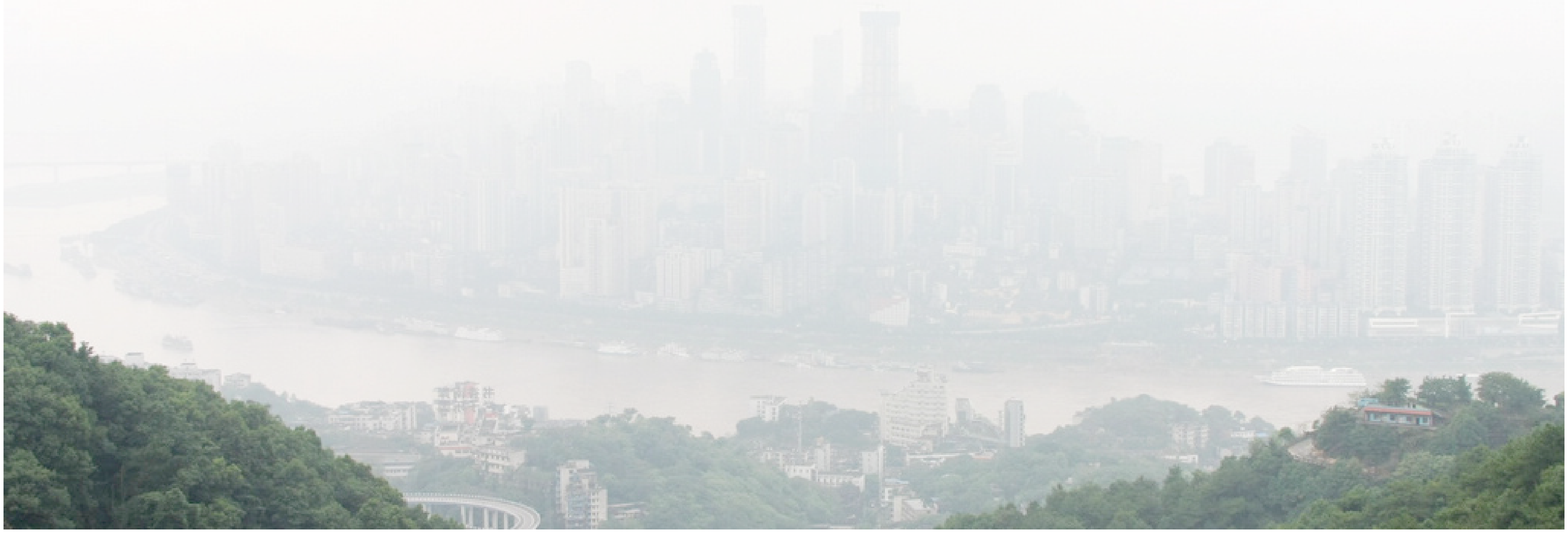} \\
(a) \\
\includegraphics[width=\linewidth]{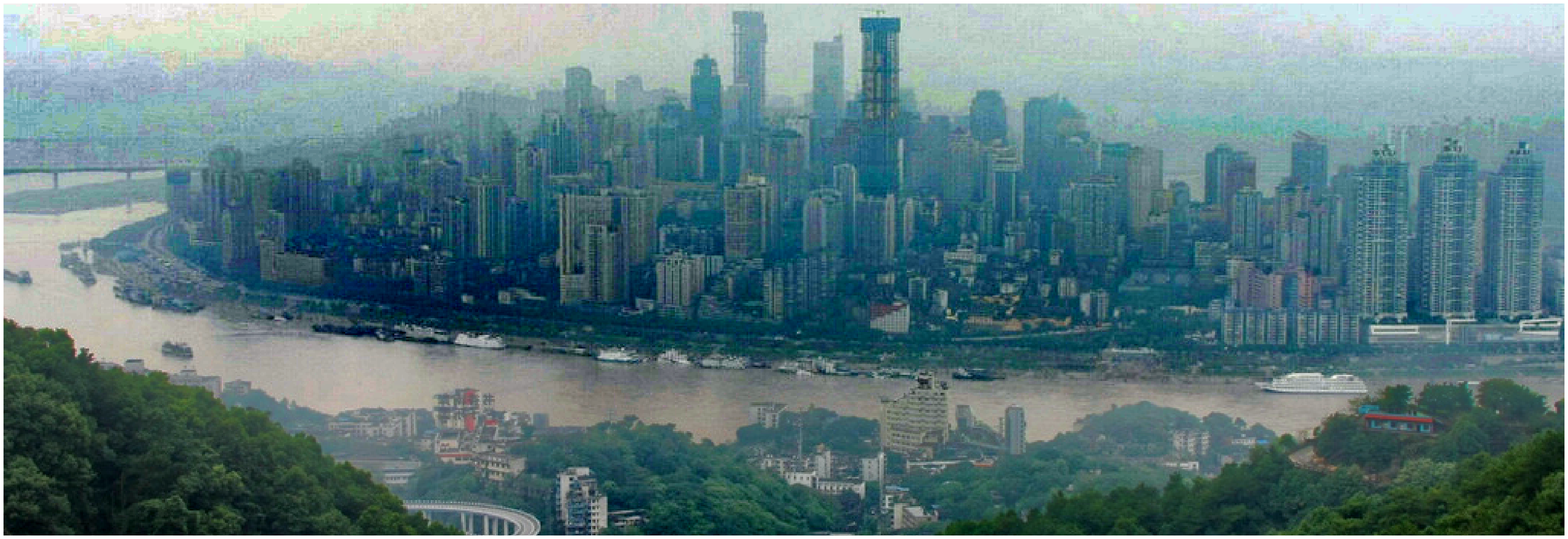} \\
(b) \\
\includegraphics[width=\linewidth]{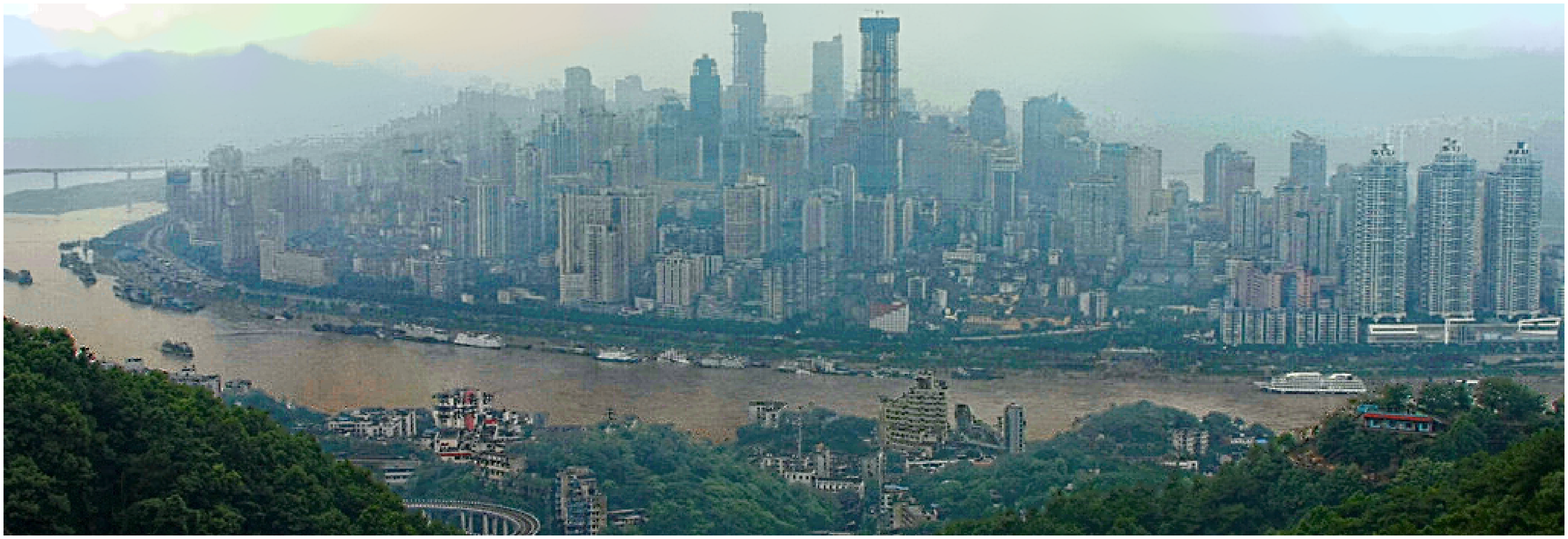} \\
(c)
\caption{ Noise in the input haze image (a) is remarkably amplified with state-of-the-art dehazing method \cite{HeST09}, yielding visually unpleasing result (b). Our result shown in (c) contains clearer image structure.}
\label{fig:compare_our}
\end{figure}

\section{Background}

Dense scattering medium is one of the main causes of inconsistency in visual perception and human understanding \cite{day2004driving}. Scattering layers often make the originally clear landmarks look distant, which explains why people think they move slower than normally when driving in fog and swimming underwater. The reduced visibility inevitably handicaps visual recognition and understanding. In contrast to its practical importance, previous approaches \cite{Schechner03,Tan08,Fattal08,HeST09,Tarel_iccv09,nishino2011bayesian} assume thin scattering layers caused, for example, by haze.

When tackling the more challenging dense-scattering-medium problem where visibility is significantly reduced, we notice inherent issues. On the one hand, the underlying structure is contaminated, making its restoration require spatially neighboring information. On the other hand, ubiquitous camera noise, image artifacts, and physically existing impurities (such as dust) in the media could be greatly amplified and influential in visual restoration. If they are not dealt with properly, erroneous estimation of structures shrouded by dense scattering media could be resulted in. It is noteworthy that intuitively applying denoising \cite{dabov2008image,MairalBPSZ09} or performing regularized inversion \cite{SchechnerA07} is not competent to solve this issue, due to the spatially varying properties.

We tackle the vital structure preserving and noise suppressing issues by proposing several novel strategies to properly enhance pictures shot in fog, haze, dust, and even underwater scenes. Both degraded structure recovery and significant noise suppression lead to the use of neighboring information, which motivates the use of non-local total variation strategy to regularize transmission and latent image estimation. It enables us to deal with noisy input, in the meantime preserving sharp discontinuities. The direct involvement of non-local terms, however, results in a complicated constrained optimization problem and is computationally very expensive. We propose a novel {\it selective-neighbor} criterion to convert it to an unconstrained continuous optimization procedure. By incorporating transmission-aware noise-control terms into the regularized energy function, the proposed method becomes very effective in dense scattering layer removal. %

\subsection{Related Work}
Central to visual restoration from scattering media is transmission estimation. On the hardware side, polarizers were used during picture taking, which help estimate part of the medium transitivity \cite{Schechner03} or augments visibility for underwater vision~\cite{SchechnerK04}. 3D scene models were used in \cite{KopfNCCCDUL08} to guide transmission estimation.

Single-image software solutions are also popular \cite{NayarN99,NarasimhanN03,Tan08,Fattal08,HeST09}. They are generally based on priors on transmission and scene radiance. Tan \cite{Tan08} developed a method mainly based on the observation that images with enhanced visibility have higher contrast and airlight depends on the distance to the viewer. Fattal \cite{Fattal08} regarded transmission and surface shading (reflection) as locally uncorrelated in a hazed image. Independent Component Analysis (ICA) was employed to estimate scene albedo and medium transitivity. A dark-channel prior was proposed in \cite{HeST09} to initialize transmission estimation followed by refinement through soft matting.

These methods produce the results by simply inverting transmission blending with the underlying structures, which can generally magnify image noise and visual artifacts. One example is shown in Fig. \ref{fig:compare_our}, where image shown in (b), the result by direct inversion, becomes noisy after haze-removal. This is the major problem when dealing with dense media, where ubiquitous floating impurities can be notably intensified.

Regularized inversion was employed in \cite{SchechnerA07}. Local regularization however by nature cannot handle strong image noise and compression artifacts. Taral {et al.} \cite{Tarel_iccv09} used depth dependent median filtering. In dense media, the sizes are hard to determine accurately. The point-wise transmission constraint was recently introduced in \cite{nishino2011bayesian}, accompanying with an Expectation-Maximization (EM) solver. It does not tackle the noise boosting problem. Joshi {et al.} \cite{Removal_seeingmt10} explicitly pointed out the noise issue and developed a method to remove artifacts by averaging {\it multiple} images with weights. In comparison, our method is a robust single-image approach taking into account the noise suppression, transmission estimation, and computation efficiency. A unified framework is developed for enhancing pictures taken in challenging underwater environment or in meteorological phenomena.

\begin{figure}
\centering
\begin{tabular}{@{\hspace{0.0mm}}c@{\hspace{2mm}}c@{\hspace{0.0mm}}}
  \includegraphics[width=0.48\linewidth]{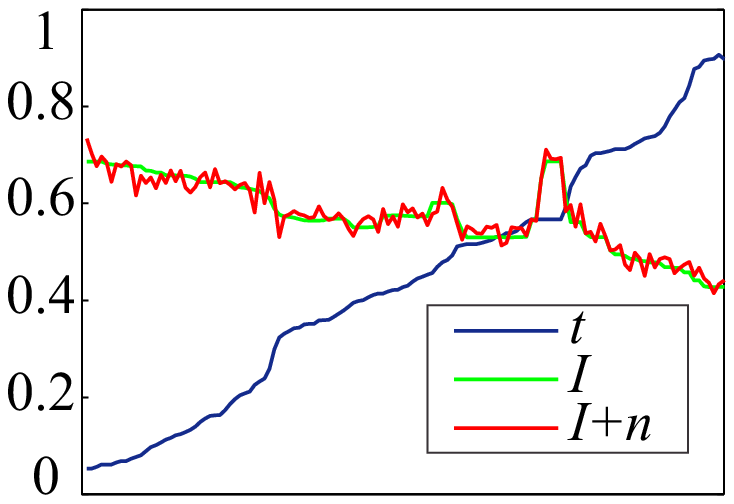} &
  \includegraphics[width=0.49\linewidth]{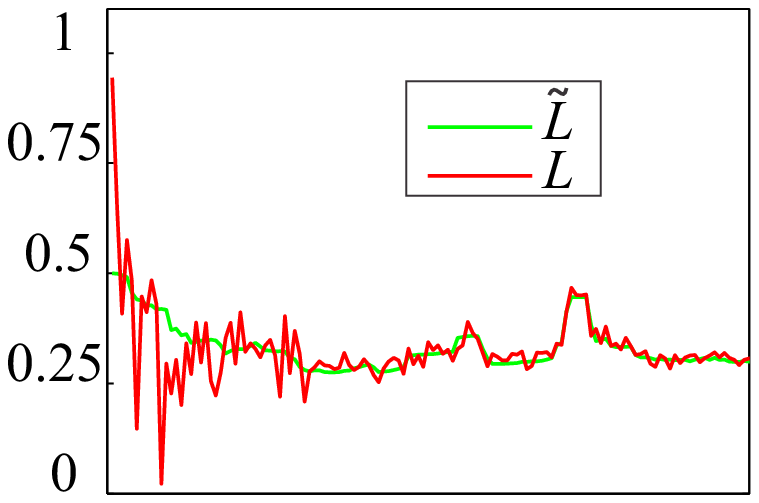}\\
   (a) &  (b)
\end{tabular}
\caption{ Noise magnification. 1D signals $I$ and $I+n$ in (a) are with very small difference due to noise (x-axis: position; y-axis: value). However for the part with small transmission $t$ (left side of (a)-(b)), the finally computed $L$ in (b) is notably dissimilar to the ground truth $\tilde{L}$ owing to noise amplification.  } \label{fig:noise1d}
\end{figure}

\subsection{The Problem}
The model of surface radiance blended with the backscattered light can be simply expressed as
\begin{eqnarray}
I({\mathrm{x}})=t({\mathrm{x}})L({\mathrm{x}})+(1-t({\mathrm{x}}))B,
\label{eq:model1}
\end{eqnarray}
where $L({\mathrm{x}})$ denotes the surface radiance that we would have sensed without the scattering medium. ${\mathrm{x}}$ indexes the 2D coordinates. $B$ is the backscattered light color vector determined by ambient illumination, also referred to as airlight or veiling light. $t({\mathrm{x}})$ is the transmission component which relates to the scene depth $d({\mathrm{x}})$ through $e^{-\eta d({\mathrm{x}})}$, $\eta$ is the attenuation coefficient, determined by scattering property of the medium.

Scattering layer removal requires an estimation of the transmission map $t({\mathrm{x}})$, the light vector $B$, and then more importantly, the restoration the latent image $L({\mathrm{x}})$. Based on the estimation of $t$ and $B$, the latent image $L$ can be recovered as
\begin{eqnarray}
L(\mathrm{x})=B-\frac{B-I(\mathrm{x})}{t(\mathrm{x})}.
\label{eq:reconstruct}
\end{eqnarray}
The simple inversion works well for general thin scattering media \cite{Schechner03,Fattal08,HeST09}. It however invokes problem for pixels with small transmission $t$, which happens when the object is distant or the medium is dense.

We now analyze how the estimation errors $\Delta t$ would affect the result. In Eq. (\ref{eq:reconstruct}), $\Delta t$ causes the error of $L$. We estimate its magnitude as
\begin{align}
|\delta_L| & = & \left|(B-\frac{B-I}{t+\Delta t}) -(B-\frac{B-I}{t}) \right| \nonumber \\
& = & \left| \frac{B-I}{t} \cdot \frac{\Delta t}{t+\Delta t} \right|
\leq \left| \frac{\Delta t}{t+\Delta t} \right|.
\label{eq:deltaL}
\end{align}
It indicates that the magnitude $|\delta_L|$ does not increase quickly with $\Delta t$. In fact, the larger $\Delta t$ is, the slower it expands. $10\%$ error in $t$ would result in a similar amount of error in $L$.

In comparison, considering the inevitable camera noise, even though it is visually unnoticeable within the fog layer, its saliency could be notably raised after dehazing. Denoting the input image noise as $n$, the resulted noise magnitude $|\delta_n|$ between the finally estimated ${L}$ using Eq. (\ref{eq:reconstruct}) and the ground truth $\tilde{L}$ without noise is calculated as
\begin{align}
|\delta_n| = \left|(B-\frac{B-(I+n)}{t}) -
(B-\frac{B-I}{t}) \right|=\left|\frac{n}{t}\right|.
\label{eq:deltan}
\end{align}
For pixels with the corresponding $t$ smaller than $0.1$, which happens when the object is distant or the medium is dense, the resulted noise will be magnified by more than $10$ times. It reveals the fact that effective and precise noise suppression is imperative in these cases and should be regarded as similarly important as transmission estimation. Methods of \cite{Fattal08,HeST09} truncate $t$ to prevent small values, which could leave part of the haze unremoved in dense media.

Fig. \ref{fig:noise1d} illustrates the noise magnification problem for small $t$. In the plot (a), small perturbation in $I$ results large deviation in $L$ when transmission $t$ is mall, as shown in (b).

\section{Approach}

Our method consists of two major steps to respectively update the transmission layer and the latent image. We automatically detect the brightest pixels, or allow users to draw scribbles containing a few sample pixels, to determine the backscattering light $B$.

\subsection{Modeling Transmission $t$}\label{sec:transmission}
By defining the logarithmic transmission, $D(\mathrm{x}) = \ln t(\mathrm{x}) = -\eta d(\mathrm{x})$, we alter our goal to compute $D$ instead. $t$ can be afterwards calculated using $t(\mathrm{x})=e^{D(\mathrm{x})}$. The reason to define $D$ is that depth $d$ is exactly negative of $D$ for each pixel. Most natural scene priors, such as the piece-wise spatial smoothness, can be imposed on depth (and correspondingly $D$), but not on the transmission variables that form different distributions.

Re-arranging terms in Eq. (\ref{eq:model1}) and taking the logarithm yield
\begin{eqnarray}
D({\mathrm{x}})=\ln(|B-I({\mathrm{x}})|)-\ln(|B-L(\mathrm{x})|).
\end{eqnarray}
By further denoting $\bar{i}(\mathrm{x})=\ln(|B-I(\mathrm{x})|)$, and $\bar{l}(\mathrm{x})=\ln(|B-L(\mathrm{x})|)$, we express the log likelihood (without normalization) as
\begin{eqnarray}
E_D(D)=\sum_{{\mathrm{x}}}\sum_c|D(\mathrm{x})-(\bar{i}^c(\mathrm{x})-\bar{l}^c(\mathrm{x}))|^2.
\label{eq:ed}
\end{eqnarray}
To model the smoothness property of the scene depth and at the same time preserve discontinuities, we resort to a non-local total variation regularizer. Our model considers neighboring pixels as well as non-neighboring ones in local windows, which is expressed as
\begin{eqnarray}
E_S(D)=\sum_{\mathrm{x}}\sum_{\mathrm{y}\in W(\mathrm{x})}
w_d(\mathrm{x},\mathrm{y})|D({\mathrm{x}})-D({\mathrm{y}})|,
\label{eq:es}
\end{eqnarray}
where $W(\mathrm{x})$ is a patch centered at $\mathrm{x}$. $w_d(\mathrm{x},\mathrm{y})$ plays the key role in shaping the local support. It is weak when $D({\mathrm{x}})$ and $D({\mathrm{y}})$ are found not necessarily similar, and is a large weight otherwise.

\begin{figure*}
\begin{tabular}{@{\hspace{0.0mm}}c@{\hspace{0.5mm}}c@{\hspace{0.5mm}}c@{\hspace{0.5mm}}c@{\hspace{0.5mm}}c@{\hspace{0.0mm}}}
\includegraphics[width=0.195\linewidth]{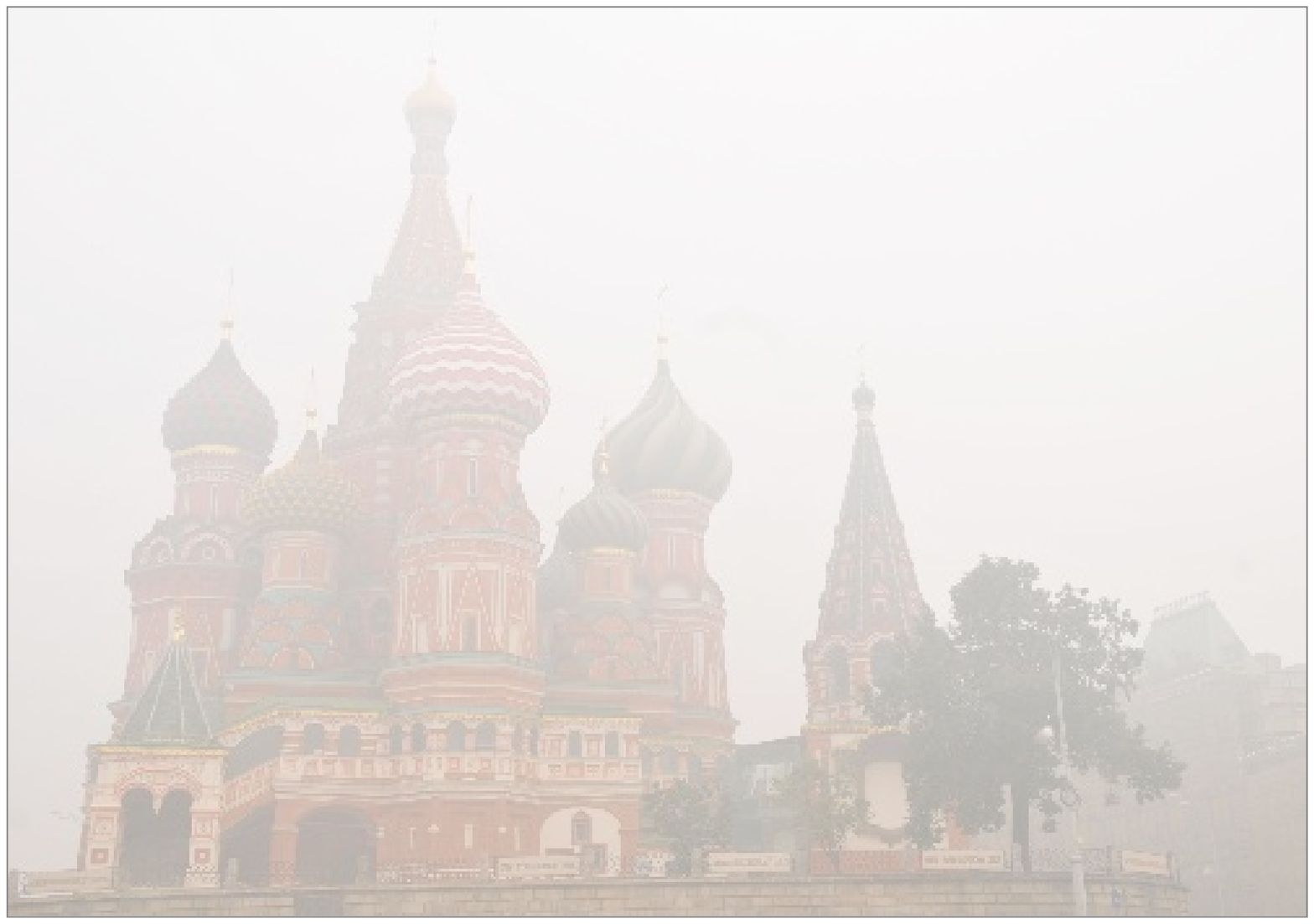}&
\includegraphics[width=0.195\linewidth]{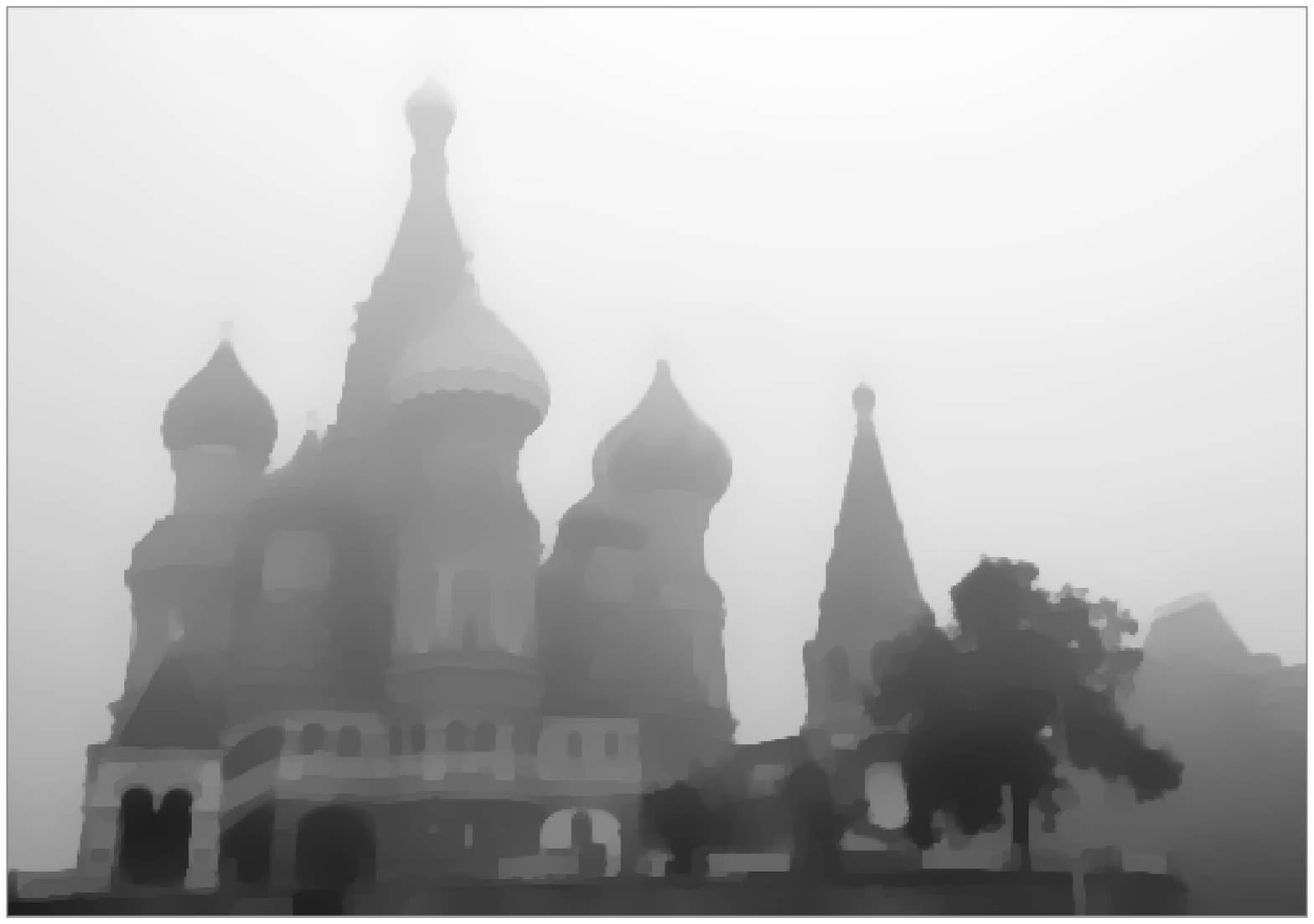}&
\includegraphics[width=0.195\linewidth]{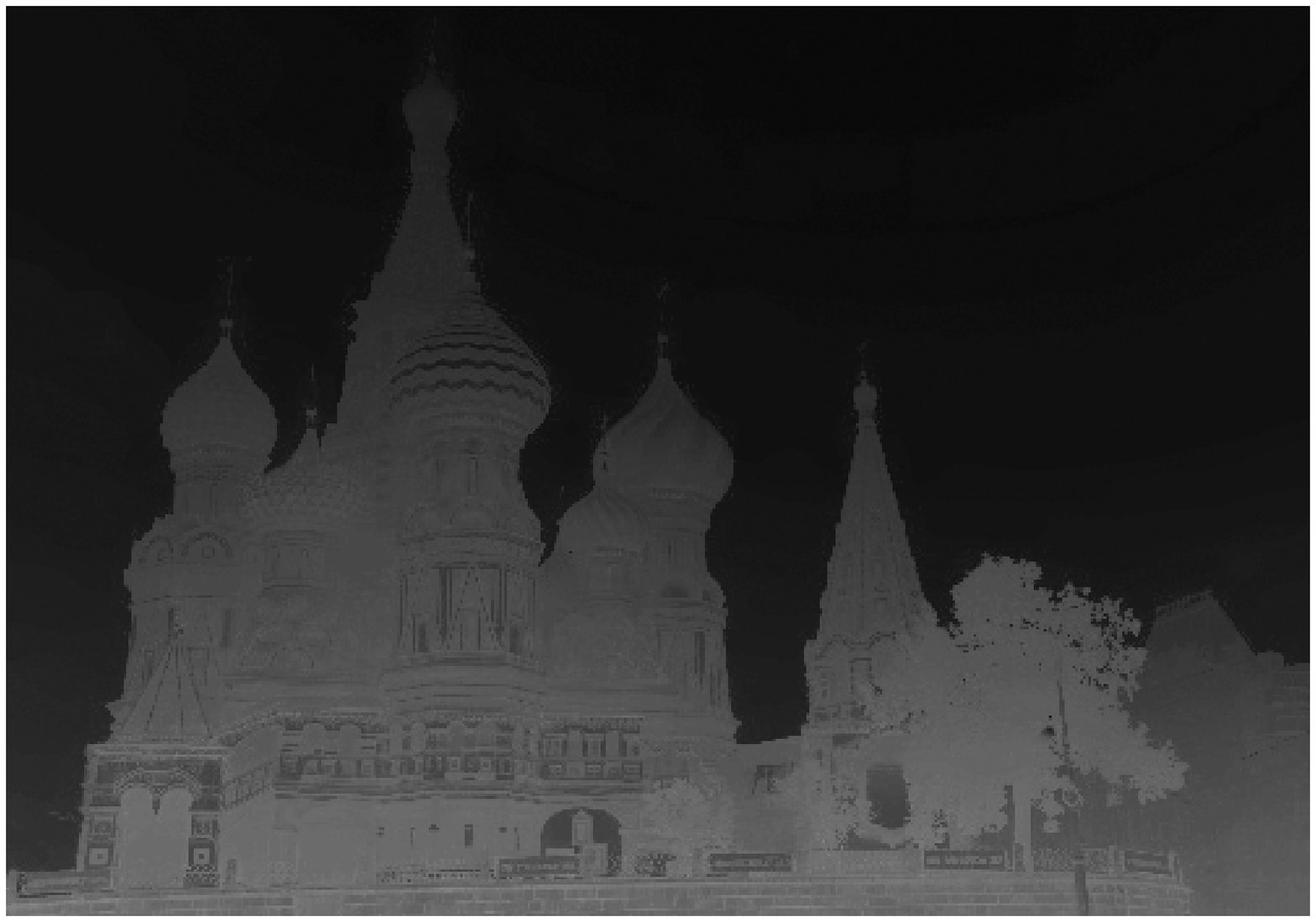}&
\includegraphics[width=0.195\linewidth]{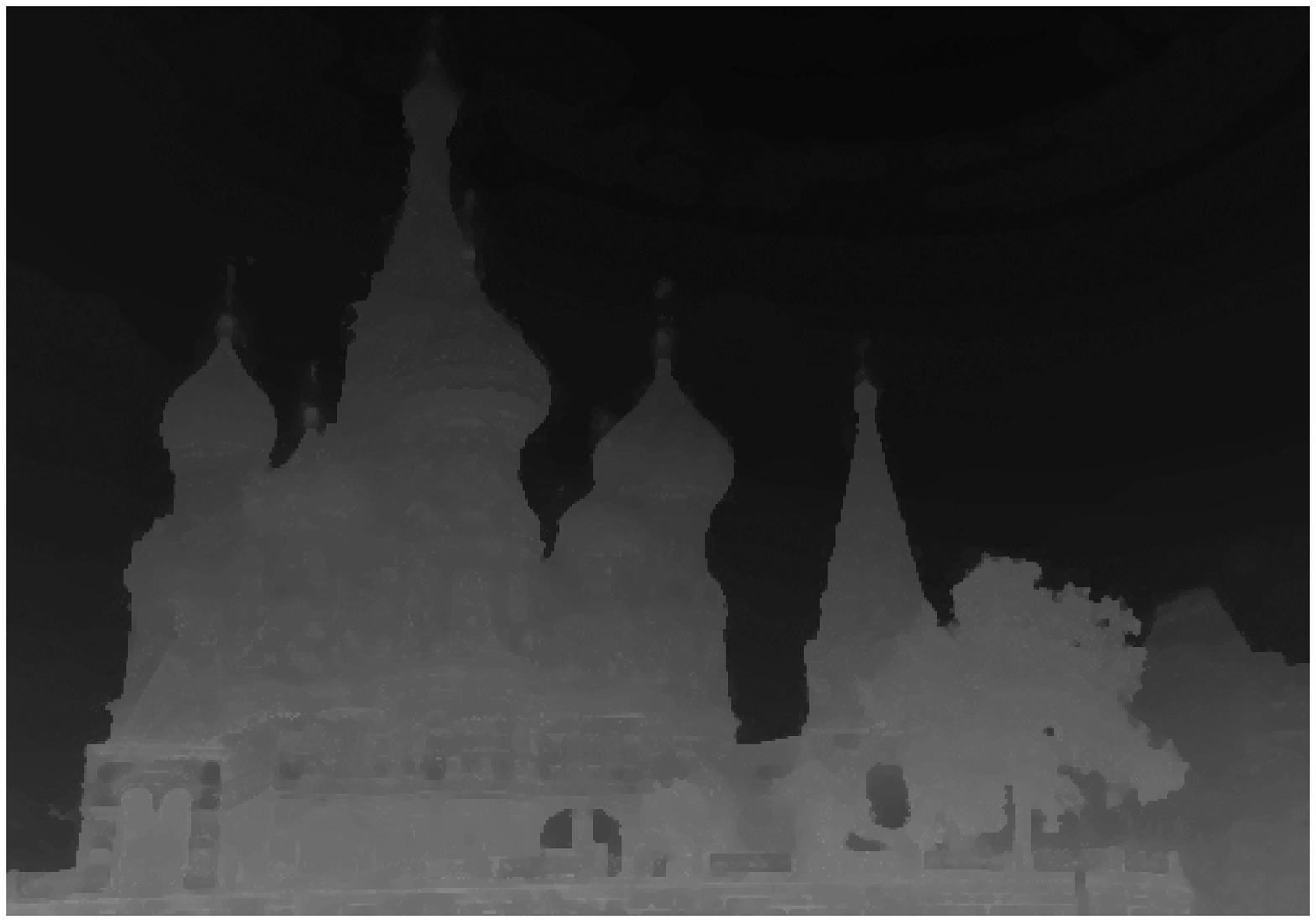}&
\includegraphics[width=0.195\linewidth]{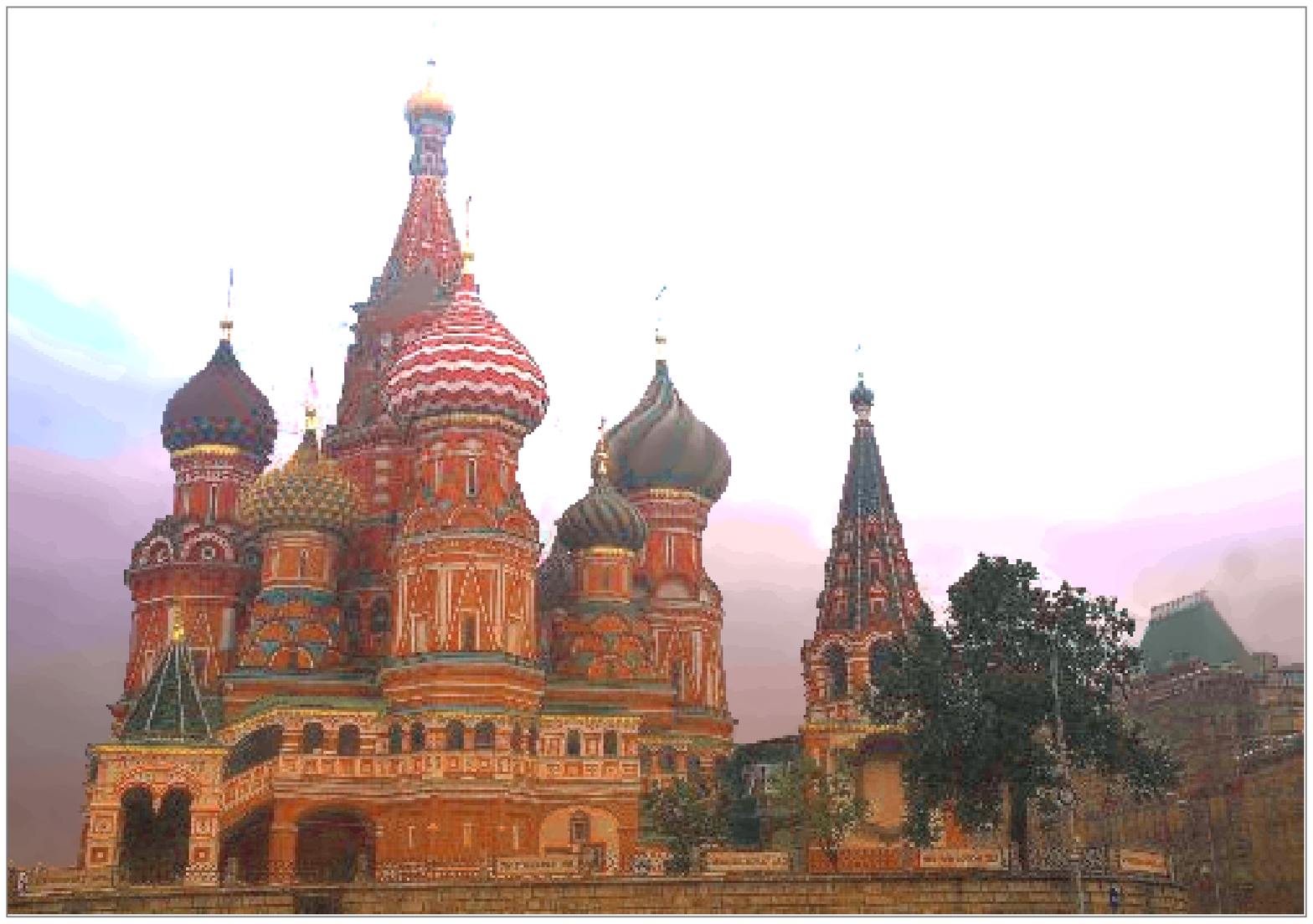}\\
{ (a) Input} & { (b) Structure map $S$} & { (c) $t$ estimated } & { (d) $t$ estimated } & { (e)  Result with $t$} \\
& & { without $S$} & {with $S$} & { in (d)}
\end{tabular}
\caption{ Structure layer extraction and its importance in transmission estimation compared to defining $w_d$ on the input image $I$.}\vspace{-0.00in} \label{fig:struct}
\end{figure*}

One may define $w_d$ in Eq. (\ref{eq:es}) with respect to the structures in the input observation $I$. That is, if two pixels $\mathrm{x}$ and $\mathrm{y}$ have similar values in $I$, $w_d(\mathrm{x},\mathrm{y})$ needs to be large to encourage strong spatial smoothness. However, we found that this strategy could adversely affect regularization. For instance, a pixel within a small texture pattern makes weights $w_d(\mathrm{x},\mathrm{y})$ small for most of $\mathrm{y}$s in $W(\mathrm{x})$, which correspondingly results in the lack of strong support in regularization. Worse, small texture-like structures in these cases generally do not correspond to actual depth discontinuity. So, instead of simply basing $w_d$ on $I$, we propose a guided weight
\begin{equation}
w_d(\mathrm{x},\mathrm{y})={g(|S(\mathrm{x})-S(\mathrm{y})|,\sigma_s)},
\end{equation}
where $g(x,\sigma)=\exp(-x^2/2\sigma^2)$. $S$ is a structure map produced using the texture-structure decomposition method \cite{tsmoothing2012} to suppress the excessive details. One example is shown in Fig. \ref{fig:struct}. The structure map is useful to remove excessive details while still preserving large-scale structures.

\paragraph{Constrained Model}

There is a lower-bound for pixel-wise transmission, based on the non-negativity of scene luminance. It always holds that
\begin{eqnarray}\label{eq:dark1}
t({\mathrm{x}})= \frac{B-I({\mathrm{x}})}{B-L({\mathrm{x}})} \geq 1-\frac{I({\mathrm{x}})}{B}.
\end{eqnarray}
The inequality is derived from $L({\mathrm{x}})\geq 0$. Therefore, the lower-bound for transmission given that all color components of $B$ are not zero can be derived as
\begin{eqnarray}
t({\mathrm{x}})\geq \max\left(1-\min_{c\in\{r,g,b\}}\frac{I^c({\mathrm{x}})}{B^c}, 0\right).\label{eq:dark_final}
\end{eqnarray}
This property can inhibit the adverse flattening effect in computing $t$, which accordingly prevents the halo artifacts in the latent image estimation.

\begin{figure*}[t]
\centering
\begin{tabular}{@{\hspace{0.0mm}}c@{\hspace{1mm}}c@{\hspace{1mm}}c@{\hspace{1mm}}c@{\hspace{1mm}}c}
  \includegraphics[width=0.195\linewidth]{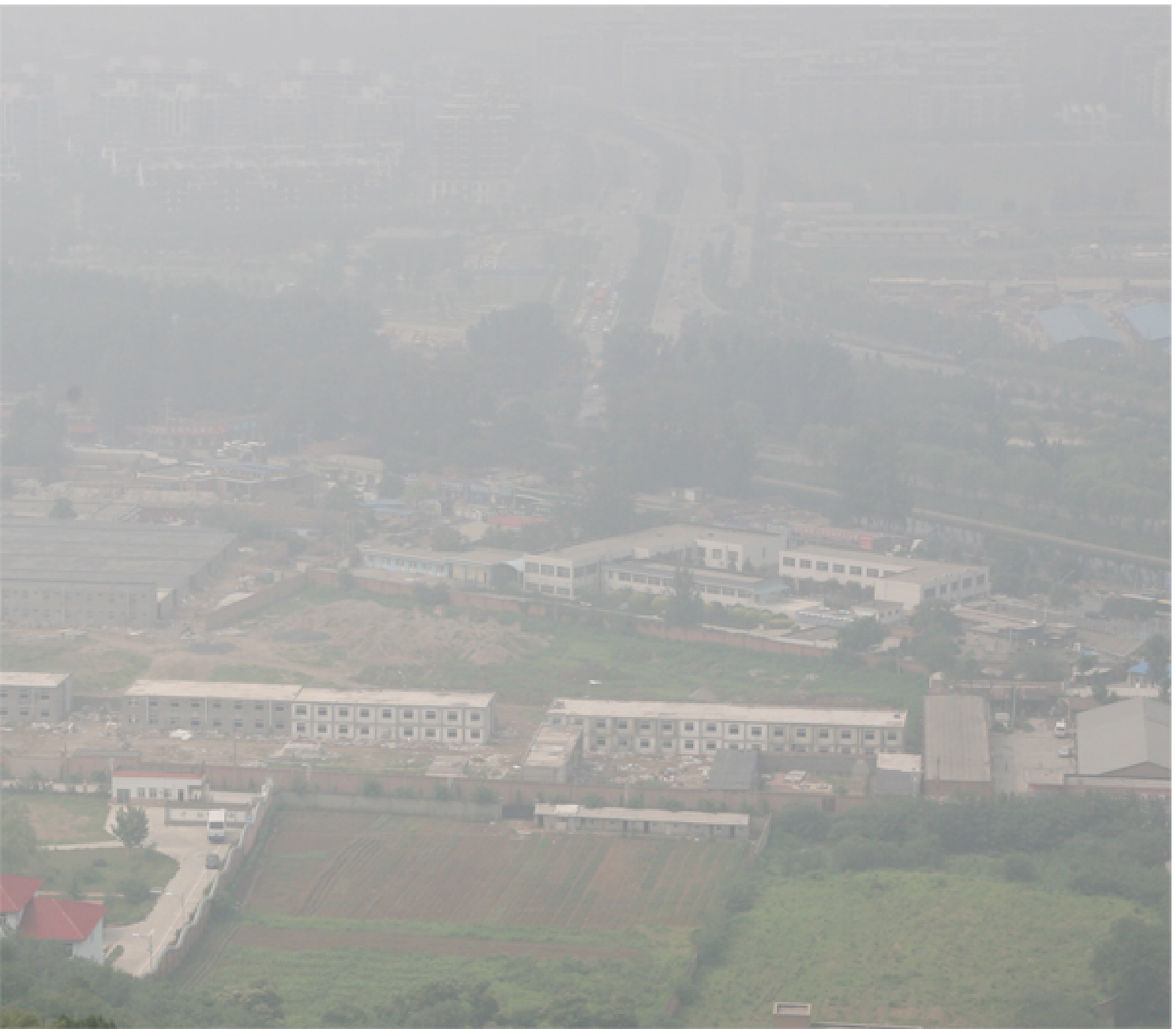}&
  \includegraphics[width=0.195\linewidth]{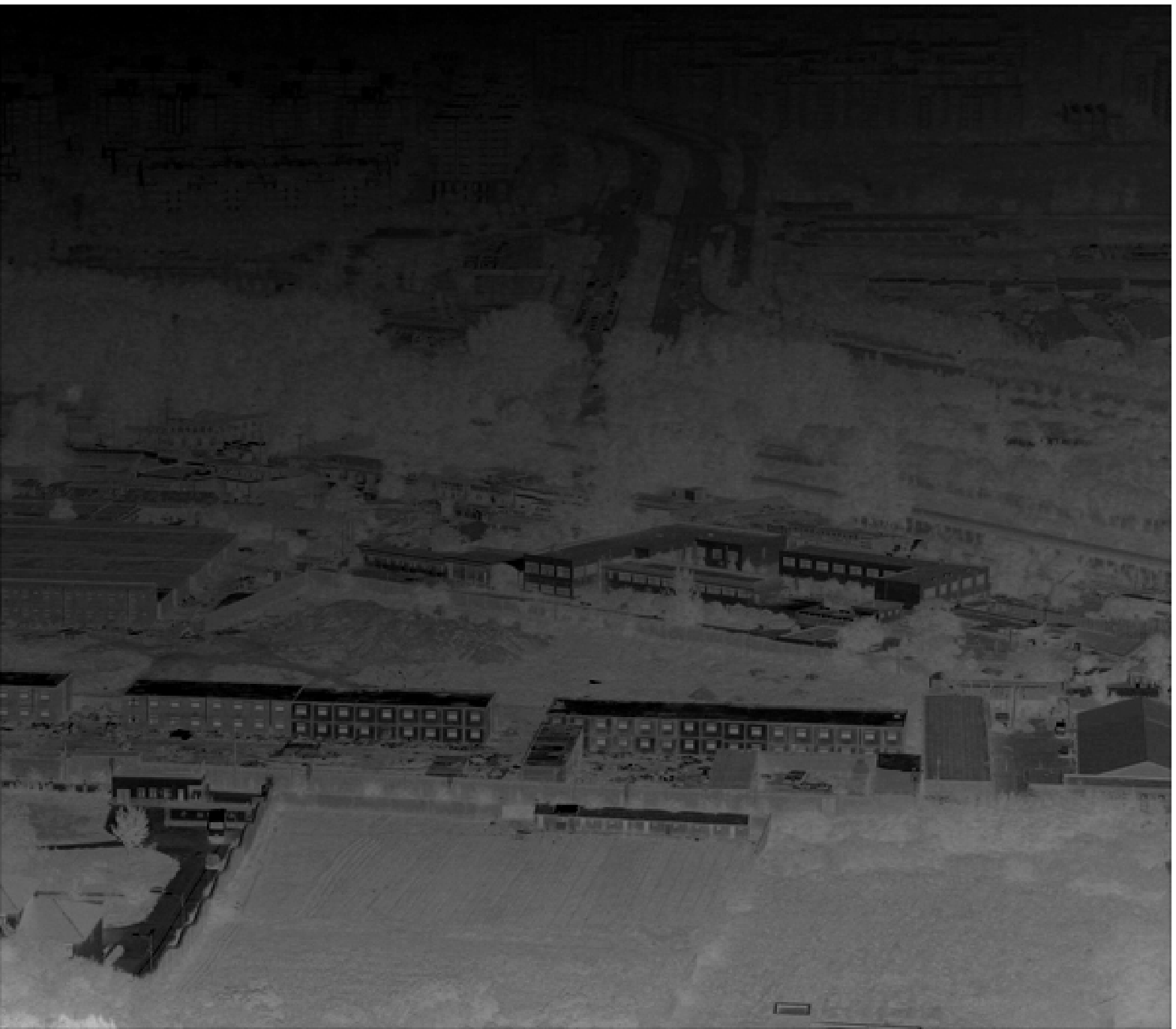}&
  \includegraphics[width=0.195\linewidth]{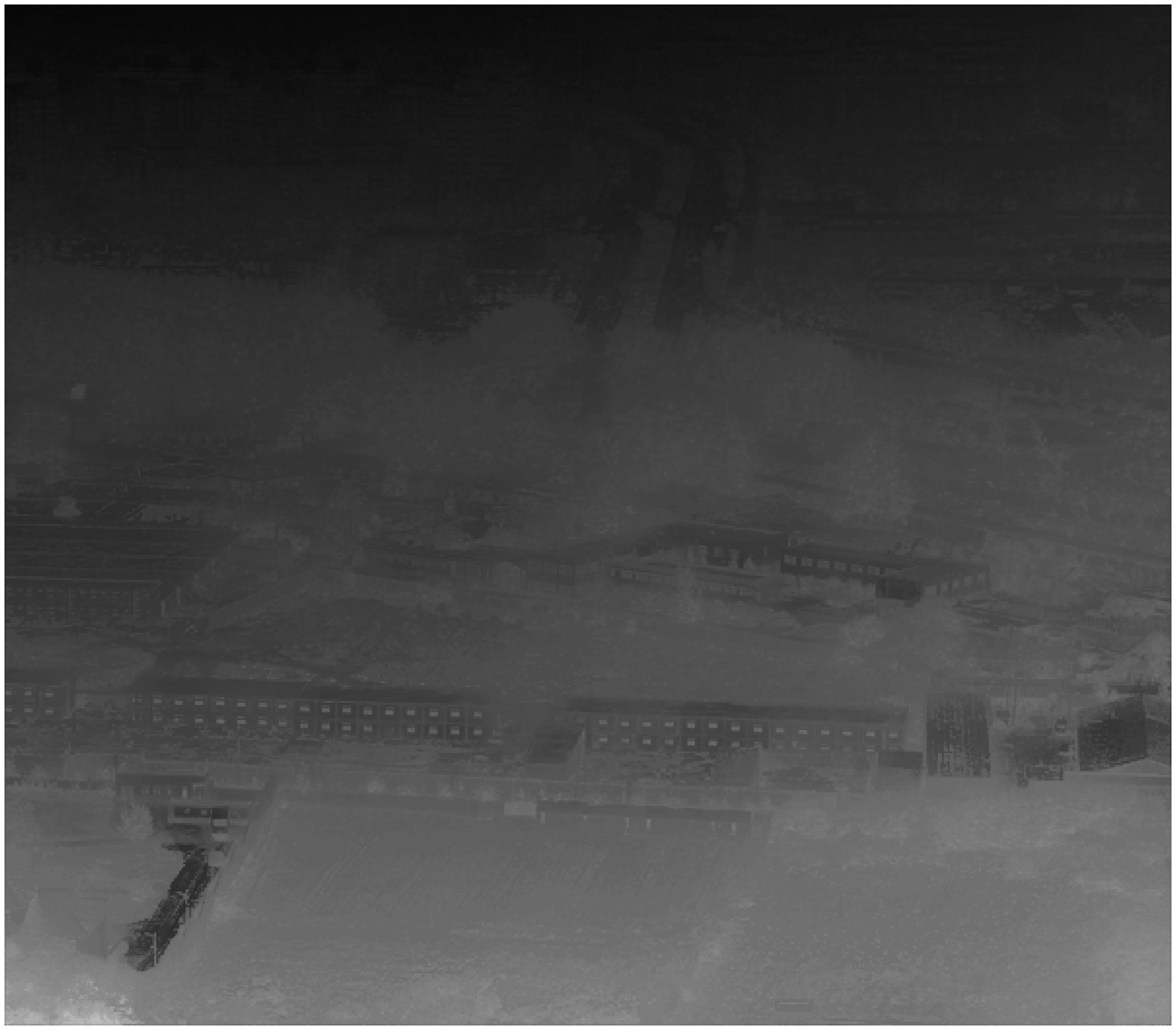} &
  \includegraphics[width=0.195\linewidth]{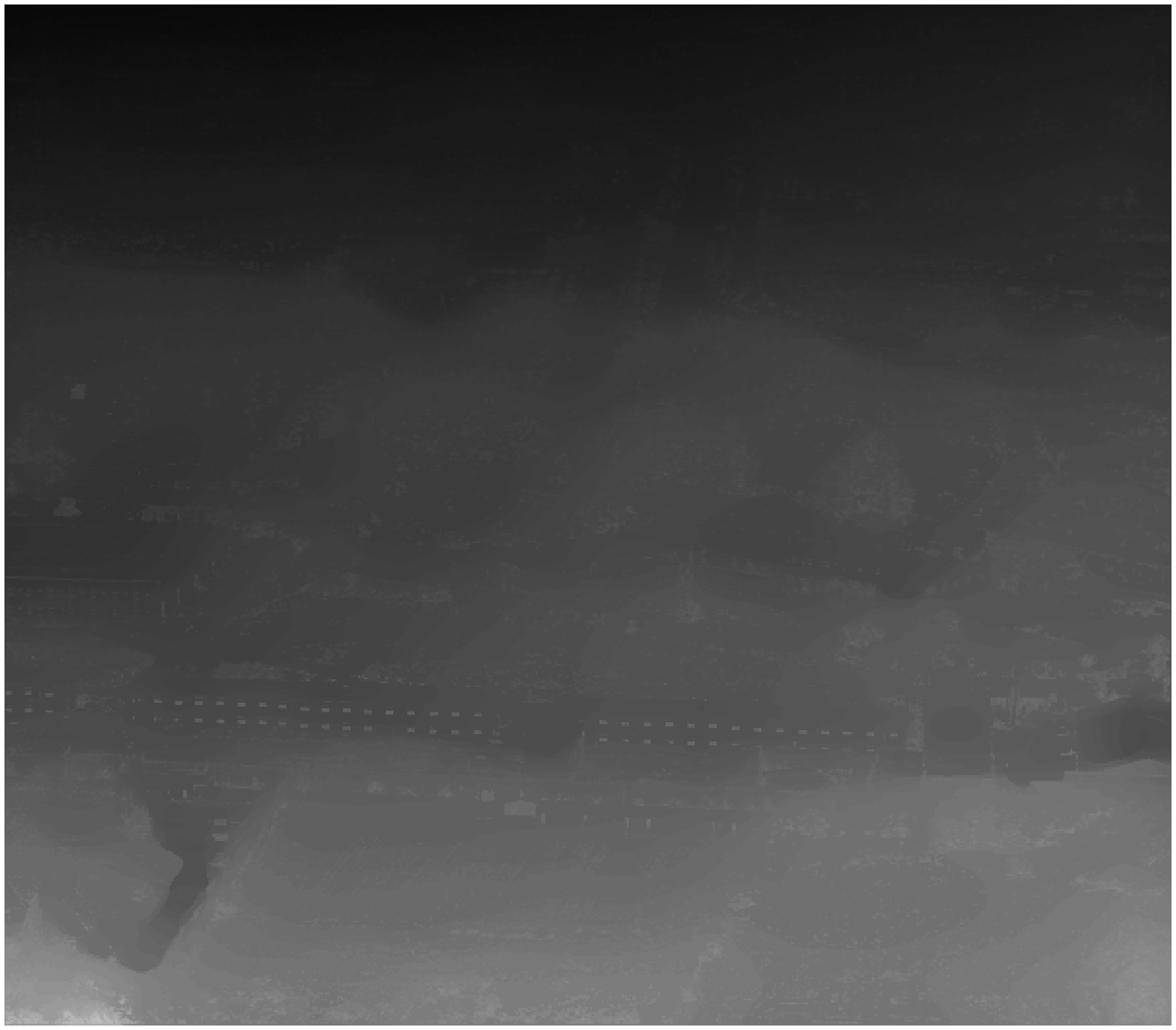} &
  \includegraphics[width=0.195\linewidth]{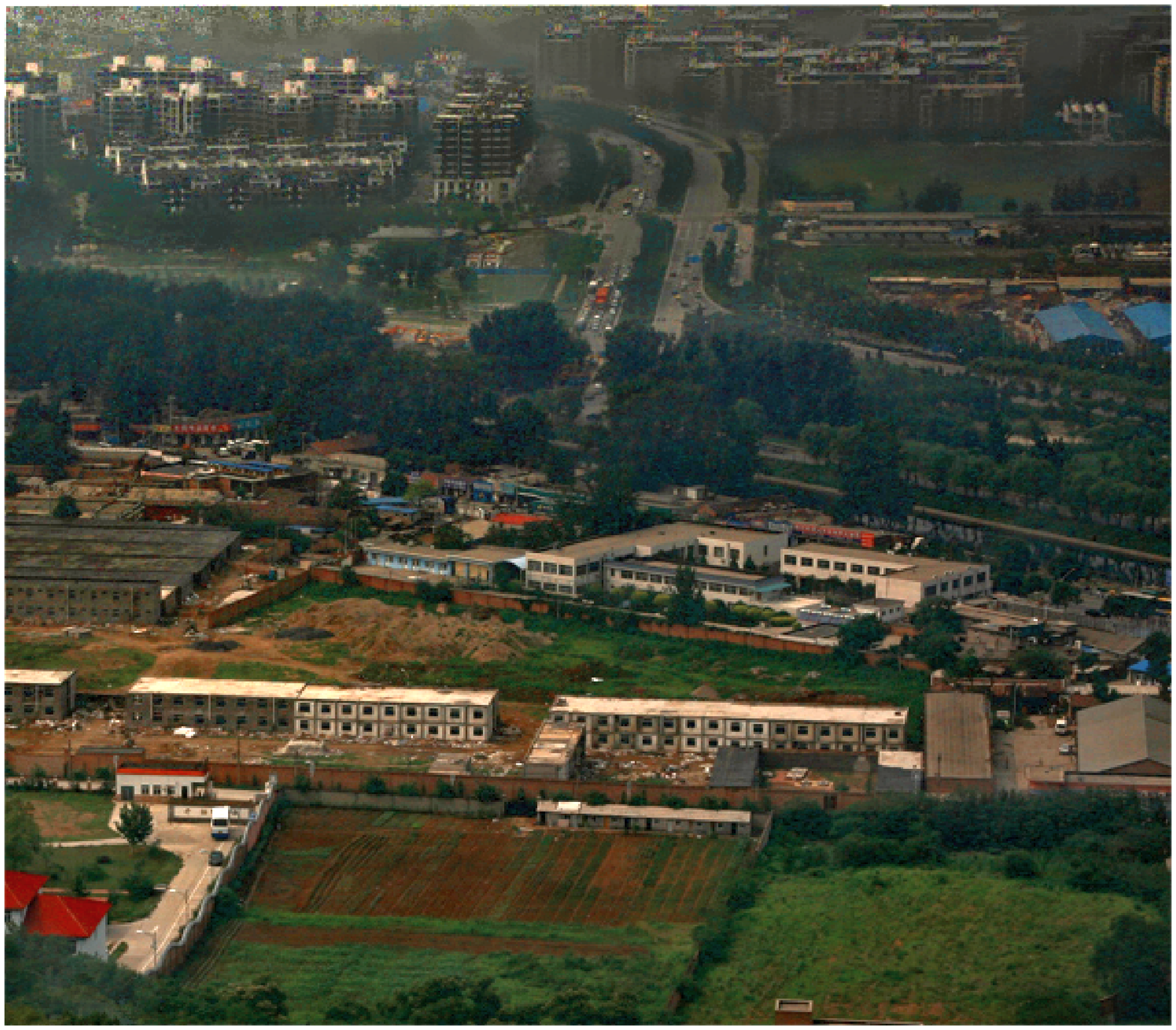} \\
  (a) Input &
  (b) Depth $D^0$ &
  (c) Depth $D^1$ &
  (d) Depth $D^2$ &
  (e) Result
\end{tabular}
\caption{ Depth maps (i.e., negative $D$ maps) estimated in two iterations. Depth map $D^0$ is initialized with the lower bound measure $v$.} \label{fig:transmission}
\end{figure*}

Now, given the terms defined in Eqs. (\ref{eq:ed}) and (\ref{eq:es}) and the constraint introduced in {Eq.} (\ref{eq:dark_final}), the final objective function to estimate $D$ (and correspondingly $t$) is written as
\begin{align}
\min\textrm{~~} & E_D+\lambda E_S  \nonumber\\
{\textrm{~~s. t.~~}} & D({\mathrm{x}})\geq v(\mathrm{x}), \label{eq:mincomplex}
\end{align}
where
\begin{equation}
v(\mathrm{x}) = \textrm{ln}\left(\max\left(1-\min_{c}\frac{I^c({\mathrm{x}})}{B^c},\epsilon\right)\right).
\label{eq:lower}
\end{equation}
$\epsilon$ is a small positive number to avoid $\ln 0$. $v$ forms a map the same size as $D$, and encodes the lower-bound of $D$ for all pixels. One example is shown in Fig. \ref{fig:transmission}(b), which specifies the smallest possible value of $D$ for each pixel.

Eq. (\ref{eq:mincomplex}) is a constrained non-linear optimization problem and is difficult to solve due to the non-local regularizer and the pixel-wise constraint. Directly applying the feasible descent direction algorithm results in very slow convergence and obtains only a local minimum even with a small-size image input.

\begin{figure}[bpt]
\centering
\begin{tabular}{c}
\includegraphics[width=0.8\linewidth]{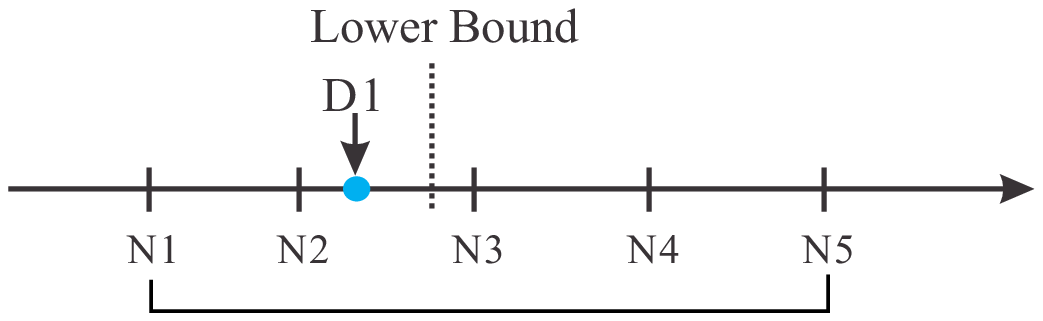}\\
(a) \vspace{0.05in}\\
\includegraphics[width=0.8\linewidth]{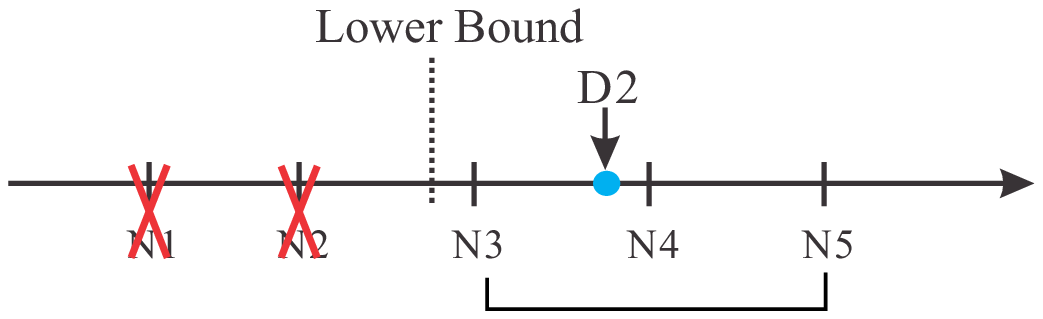}\\
(b)
\end{tabular}
 \caption{ Illustration of neighbor selection. (a) Taking all neighbors N1-N5 in regularization yields result D1, violating the transmission lower-bound condition. {(b)} Our method explicitly removes N1 and N2 in regularization because they are smaller than the lower bound, and obtain the result D2, naturally satisfying the condition.}
\label{fig:selectnn}
\end{figure}

\paragraph{Problem Conversion} We present a new method to convert the constrained optimization problem to an unconstrained one. The benefit is twofold. This conversion simplifies the objective function by embedding the inequality in the new form. Also, it enables transmission estimation by a very simple and efficient procedure. As a result, the originally difficult optimization can be nicely achieved in our method.

Our scheme is to iteratively update transmission for each pixel by computationally trackable relaxation. In each pass, we adaptively select suitable neighboring pixels $\mathrm{y}$ in Eq. (\ref{eq:es}) for regularization, based on their current transmission values,  so that Eq. (\ref{eq:dark_final}) can be explicitly satisfied. Intuitively, if one pixel $\mathrm{y}$ in Eq. (\ref{eq:es}) possibly pulls $\mathrm{x}$ out of the required range in Eq. (\ref{eq:dark_final}), we discard it in regularization. In detail, we  consider the following two cases.

{\bf 1)} When $\min_c {I^c({\mathrm{x}})}/{B^c}<1$, desirably, $D({\mathrm{x}})$ needs to satisfy the condition $D({\mathrm{x}}) \geq v({\mathrm{x}})$, where $v({\mathrm{x}})$ is the ``lower bound'' defined in Eq. (\ref{eq:lower}) and illustrated in Fig. \ref{fig:selectnn}. In regularization, we deliberately set $w_d(\mathrm{x},\mathrm{y})=0$ for all $\mathrm{y}$s, whose current transmission estimates $D(\mathrm{y}) < v({\mathrm{x}})$. So these $\mathrm{y}$s will not be used to regularize ${\mathrm{x}}$, as shown in Fig. \ref{fig:selectnn}(b). This naturally guarantees the lower-bound condition.

{\bf 2)} When $\min_c {I^c({\mathrm{x}})}/{B^c}\geq 1$, the condition $D({\mathrm{x}})\geq \ln(\epsilon)$ can always be satisfied because the current estimates $D({\mathrm{y}})$ for all $\mathrm{y}$s also satisfy the lower-bound condition, i.e., $D({\mathrm{y}})\geq \ln(\epsilon)$. Hence, Eq. (\ref{eq:es}) can be applied directly with all $\mathrm{y}$s.

In accordance to the analysis, we modify the weight definition in Eq. (\ref{eq:es}) slightly to
\begin{align}
\tilde{w}_d(\mathrm{x},\mathrm{y})=
\bigg\{\begin{array}{l@{\hspace{2mm}}l}\hspace{-0.05in}g(|S(\mathrm{x})-S(\mathrm{y})|,\sigma_s),
& \textrm{if } D(\mathrm{y})\geq v(\mathrm{x});\\
\hspace{-0.05in}0, & \textrm{otherwise}.
\end{array}\label{eq:wt_mo}
\end{align}
If $D(\mathrm{y})$ is smaller than the possible lower bound of $\mathrm{x}$, to avoid dragging $\mathrm{x}$ out of the allowed range during estimation, we break the connection between $\mathrm{x}$ and $\mathrm{y}$ by setting $\tilde{w}_n(\mathrm{x},\mathrm{y})=0$.

Normalization in each local window after removing the zeroed-out values yields
\begin{align}
w_n(\mathrm{x},\mathrm{y}) = \frac{\tilde{w}_d(\mathrm{x},\mathrm{y})}{\sum_{\textrm{y} \in W(\textrm{x}) \tilde{w}_d(\mathrm{x},\mathrm{y}) }}
\end{align}
The modified regularization term is therefore
\begin{eqnarray}
E_{\tilde{S}}(D)=\sum_{\mathrm{x}}\sum_{\mathrm{y}\in W(\mathrm{x})}
\tilde{w}_d(\mathrm{x},\mathrm{y})|D({\mathrm{x}})-D({\mathrm{y}})|.
\label{eq:es1}
\end{eqnarray}
Now without loss of generality, by incorporating the {\it selective-neighbor} scheme in regularization, we prevent  problematic estimation that violates the transmission lower-bound condition. The final objective function is written as
\begin{eqnarray}
E(D)=E_D(D)+\lambda E_{\tilde{S}}(D), \label{eq:energy_d}
\end{eqnarray}
without explicit hard constraints.

Eq. (\ref{eq:energy_d}) is not only simple in its representation, but also bears the advantage to find an extremely simple method for optimization based on computationally tractable relaxation to iteratively update $D$ for pixels respectively.

More details about the numerical solver is included in Appendix \ref{sec:est_D}.  Fig. \ref{fig:transmission} shows how the $-D$ map is improved in iterations.

\subsection{Inferring Latent Image $L$}\label{sec:latent_image}

To compute $L$ given the $t$ estimate, we do not directly solve Eq. (\ref{eq:reconstruct}) since this scheme suffers from noise magnification. Instead, we apply optimization to infer a visually plausible $L$ image.

For robustness, we define our data energy function as
\begin{eqnarray}
E_d(L)=\sum_{\mathrm{x}} t(\mathrm{x})^2 |L(\mathrm{x})-L_0(\mathrm{x})|^2, \label{eq:EDL}.
\end{eqnarray}
$L_0$ is the result intuitively computed using Eq. (\ref{eq:reconstruct}). The data energy term in Eq. (\ref{eq:EDL}) suggests that the optimized latent image should be similar to $L_0$ weighted by $t(\mathrm{x})^2$. When $t$ is large -- that is, the object is not distant -- we should trust $L_0$ because noise is not magnified too much according to our analysis.

We also provide a transmission-aware regularization term, which employs smoothness priors to further suppress noise. It is expressed as another non-local total variation:
\begin{eqnarray}
E_s(L)=\sum_{\mathrm{x}}\sum_{\mathrm{y}\in
W(\mathrm{x})}\bar{m}(\mathrm{x},\mathrm{y})|L(\mathrm{x})-L(\mathrm{y})|.
\end{eqnarray}
The weight map $\bar{m}$ is normalized from $m$ that contains two respective constraints to suppress noise. $m$ is defined as
\begin{align}
m(\mathrm{x},\mathrm{y})= & g(|t(\mathrm{x})-t(\mathrm{y})|,\sigma_t)  \nonumber\\
&~ \cdot g(\|P(L_0,\mathrm{x})-P(L_0,\mathrm{y})\|_2,\sigma_L),
\end{align}
where $g(\cdot,\sigma)$ is a Gaussian tradeoff with standard deviation $\sigma$. The first term calculates the transmission similarity between pixels, based on the fact that noise levels are magnified with respect to transmission. The second term actually measures the patch matching fidelity. $P(L_0,\mathrm{x})$ denotes a $7\times7$ window in $L_0$ centered at $\mathrm x$. $\|P(L_0,\mathrm{x})-P(L_0,\mathrm{y})\|_2$ uses a windowed L2-norm error measure to robustly estimate the color difference between pixels. Combining the two weight terms, if two pixels are in the same depth layer and have akin neighbors, their similarity is high. This patch-based error measure is much more robust than pixel-wise operations.

The final energy for estimating $L$ is therefore given by
\begin{eqnarray} E(L)=E_d(L)+\lambda_LE_s(L).
\end{eqnarray}
Our solver is similar to that for $D$ estimation. Two or three iterations are enough to produce the results. {See Appendix \ref{sec:est_L} for more details.}

\begin{figure*}[t]
\centering
\includegraphics[width=0.85\linewidth]{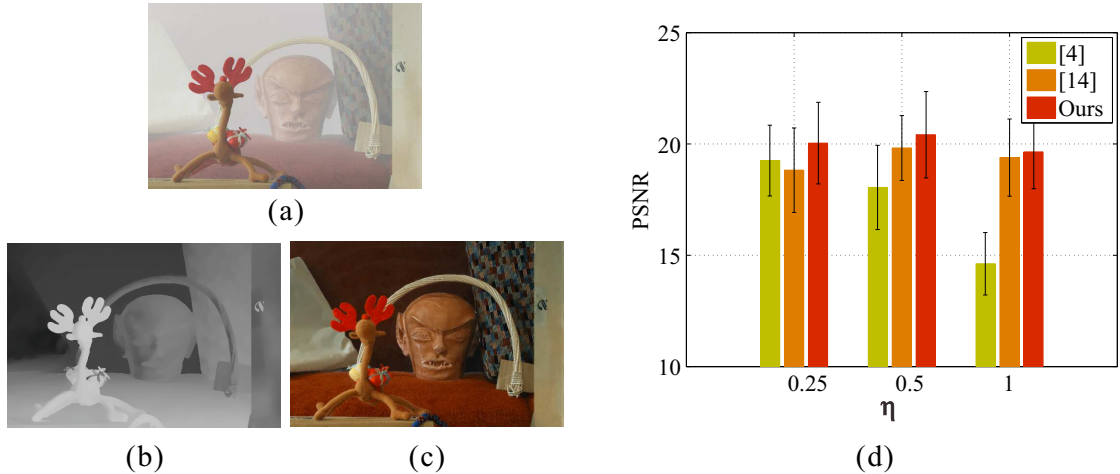}
\caption{Quantitative evaluation. (a) is the foggy input. (b) and (c) show our restored transmission and latent images. (d) shows the statistics for different approaches for three levels of fog thickness. }
\label{fig:stats}
\end{figure*}

\begin{figure*}[tb]
\centering
\begin{tabular}{@{\hspace{0.0mm}}c@{\hspace{1mm}}c@{\hspace{0mm}}}
  \includegraphics[width=0.495\linewidth]{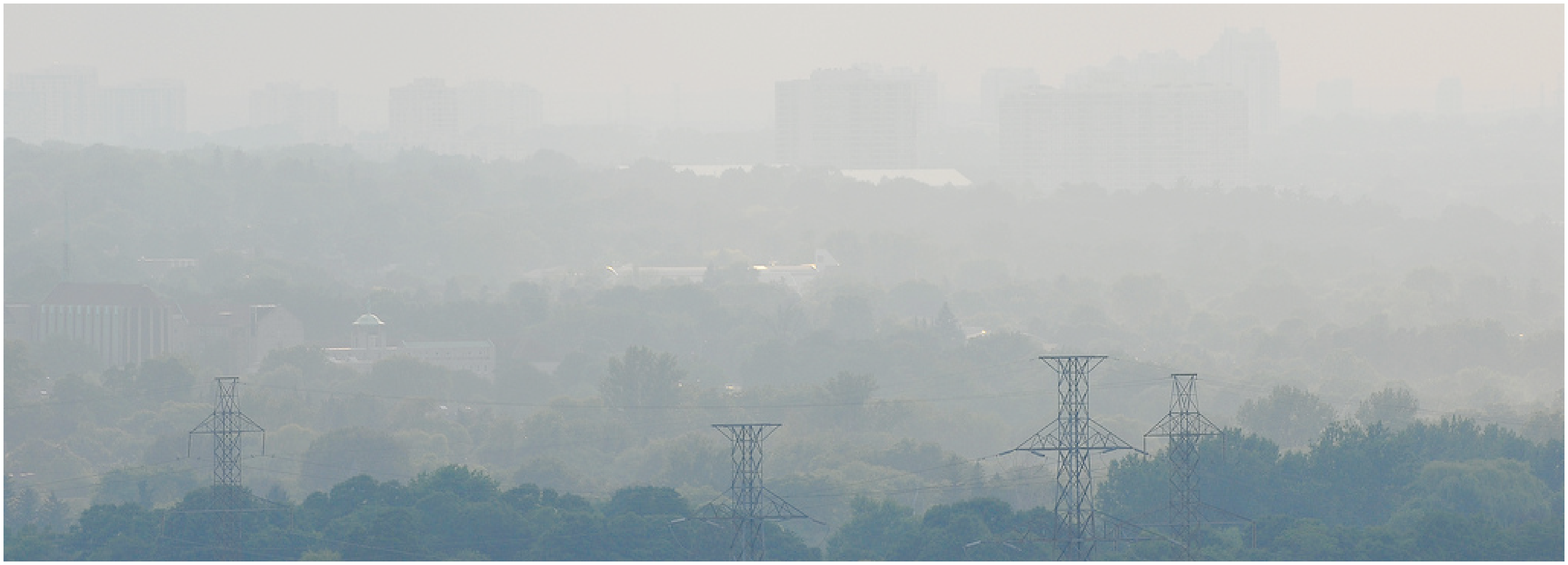} &
  \includegraphics[width=0.495\linewidth]{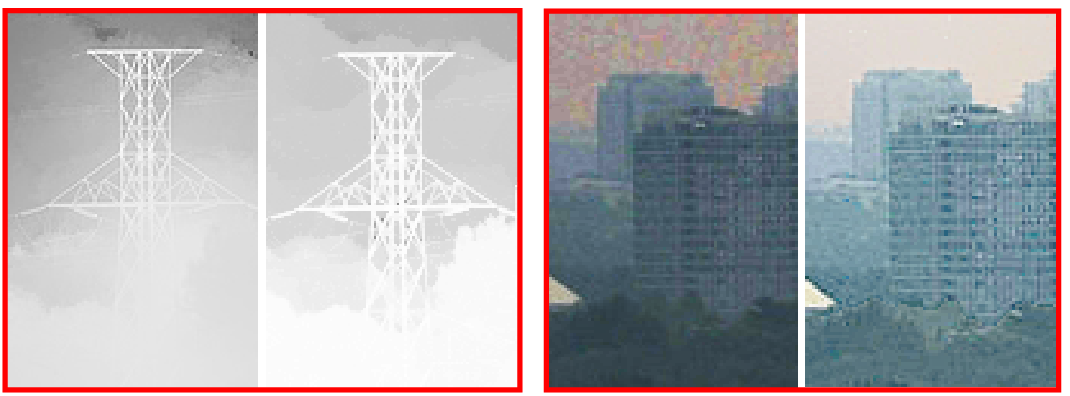} \\
  (a) &  (b) \\
  \includegraphics[width=0.495\linewidth]{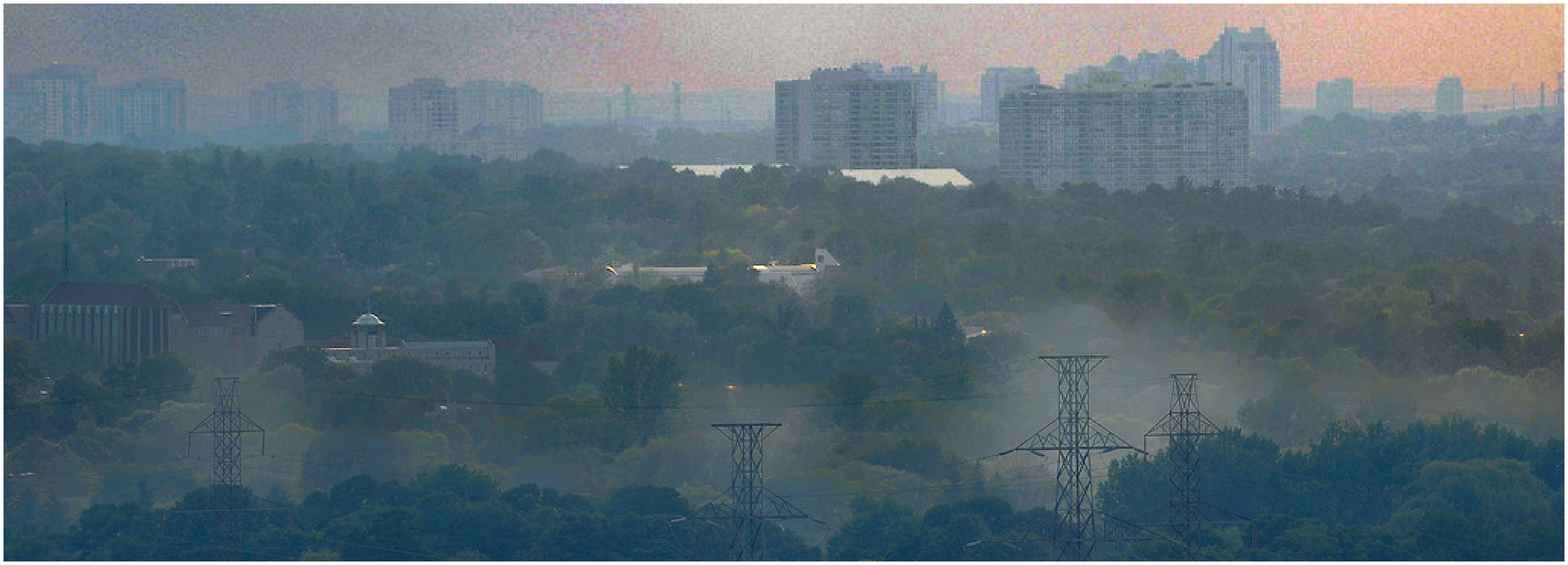}  &
  \includegraphics[width=0.495\linewidth]{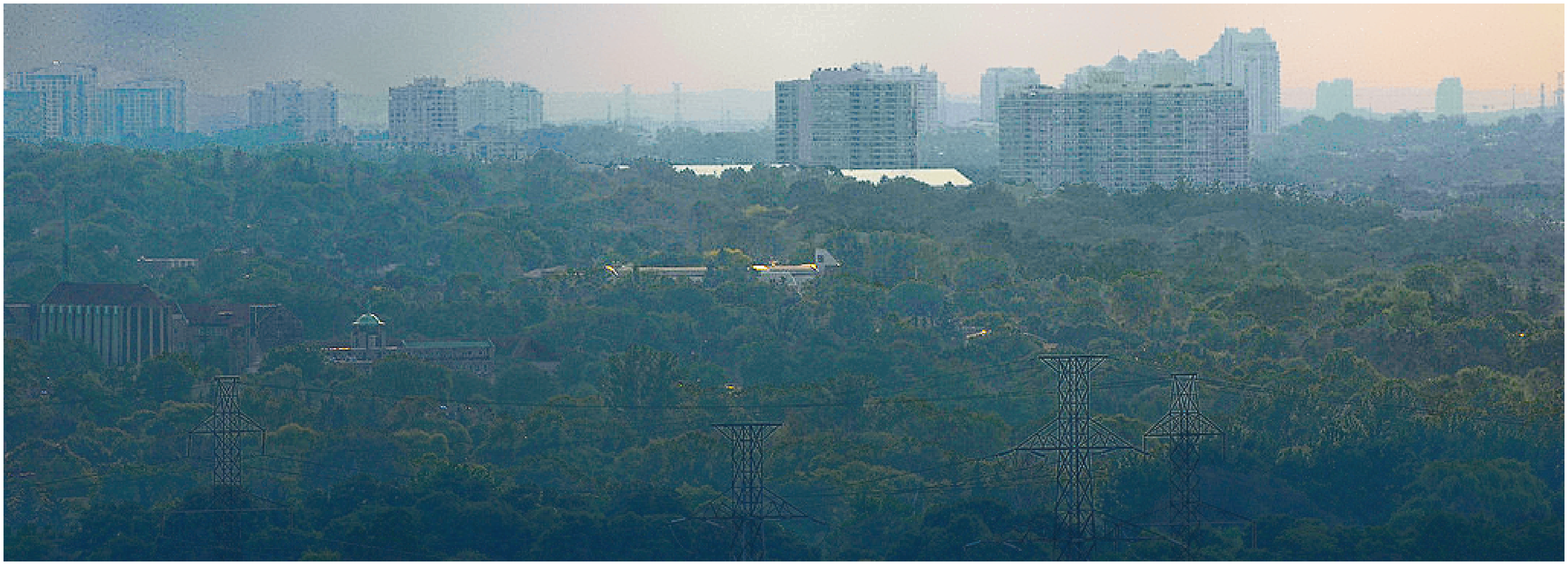} \\
   (c) &  (d)  \\
  \includegraphics[width=0.495\linewidth]{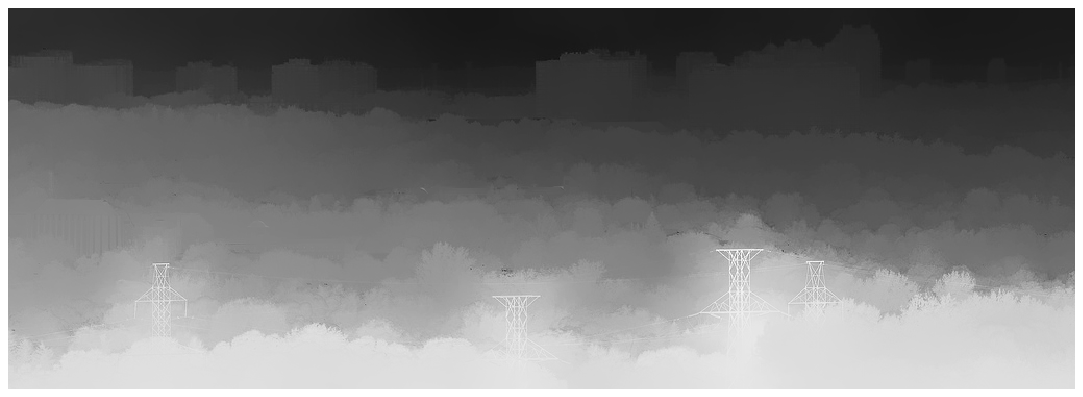}  &
  \includegraphics[width=0.495\linewidth]{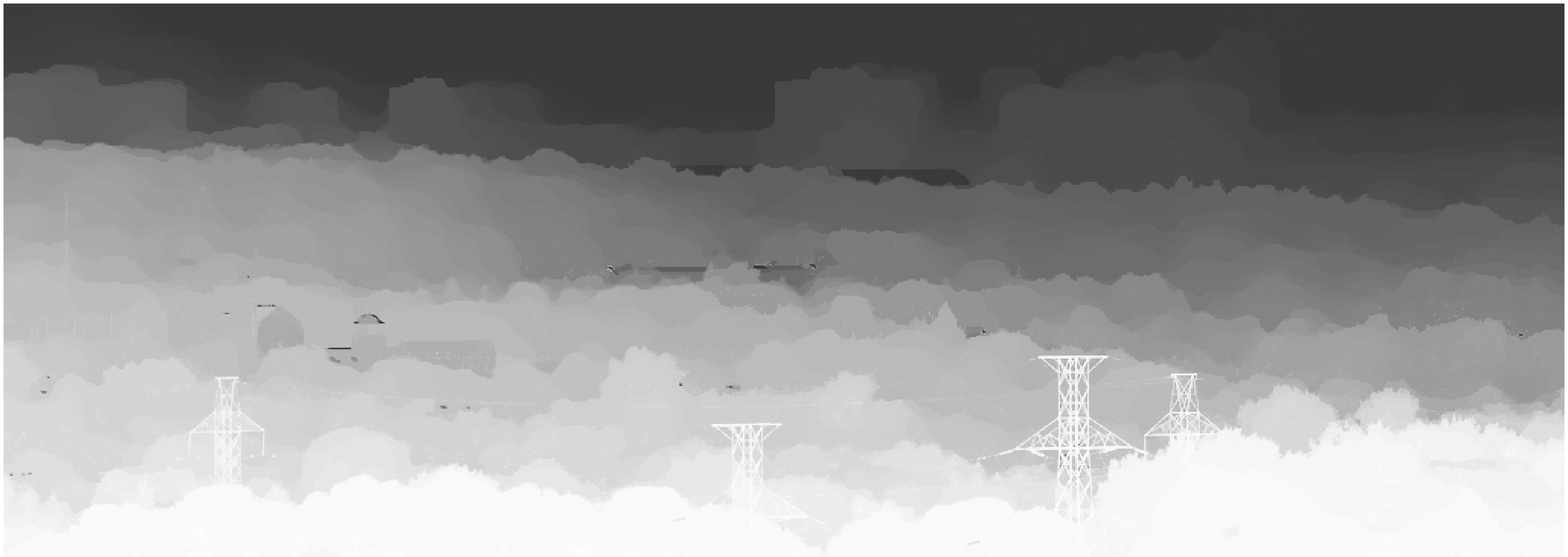} \\
   (e) &  (f)
\end{tabular}
\caption{ Fog image. (a) Input image. (c) and (e) Image $L$ and map $t$ of \cite{HeST09}, which does not consider dense scattering media. (d) and (f) Our results. (b) Close-ups. Our results are on the right. }
\label{fig:pano2}
\end{figure*}

\section{Experimental Results}\label{sec:result}

We convert the input image $I$ to a linear color space before applying our method and perform the inverse gamma correction to coarsely curtail the effect of nonlinear color transform from the camera.

\subsection{Quantitative evaluation}
We quantitatively evaluate our method using images with ground truth depth \cite{ScharsteinP07}. We collect 10 images with various color and texture patterns and generate fog contaminated inputs using Eq. (\ref{eq:model1}). Peak Signal to Noise Ratios (PSNRs) are calculated based on our restored images and the ground truth ones. Gaussian noise with $\sigma=10$ is added to the inputs to simulate noise and dust in the air. We test the performance of our method under different fog thickness settings by varying $\eta$ in $t=e^{-\eta d(\mathrm{x})}$. Sample images and the statistics are shown in Fig. \ref{fig:stats}, where (a) is an input with $\eta=1$; (b) and (c) are our restored transmission and image; (d) shows the average PSNR under different densities of the scattering medium. We compare our method with the one using dark channel prior \cite{HeST09}. To better evaluate the importance of non-local regularizer, we also compare our method with regularized haze removal \cite{SchechnerA07} by replacing non-local regularizer with a local one in the image restoration step. It can be seen that our method has a clear superiority in preserving structures with the increases of fog thickness, validating the effectiveness of our model in handling dense scattering layers. In what follows, we show more results on natural images.

\begin{figure*}[th]
\centering
\begin{tabular}{@{\hspace{0.0mm}}c@{\hspace{1mm}}c@{\hspace{0mm}}}
  \includegraphics[width=0.485\linewidth]{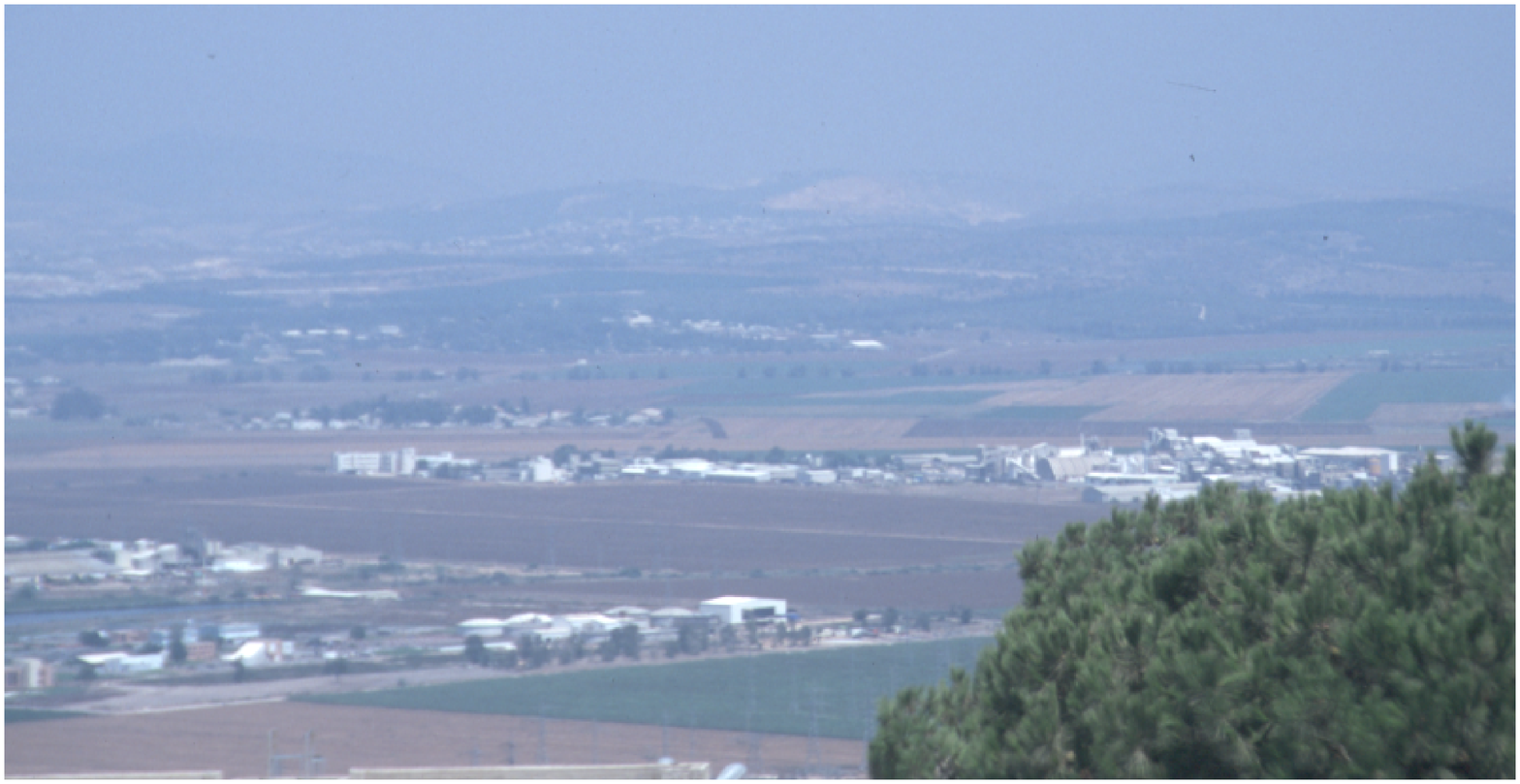} &
  \includegraphics[width=0.485\linewidth]{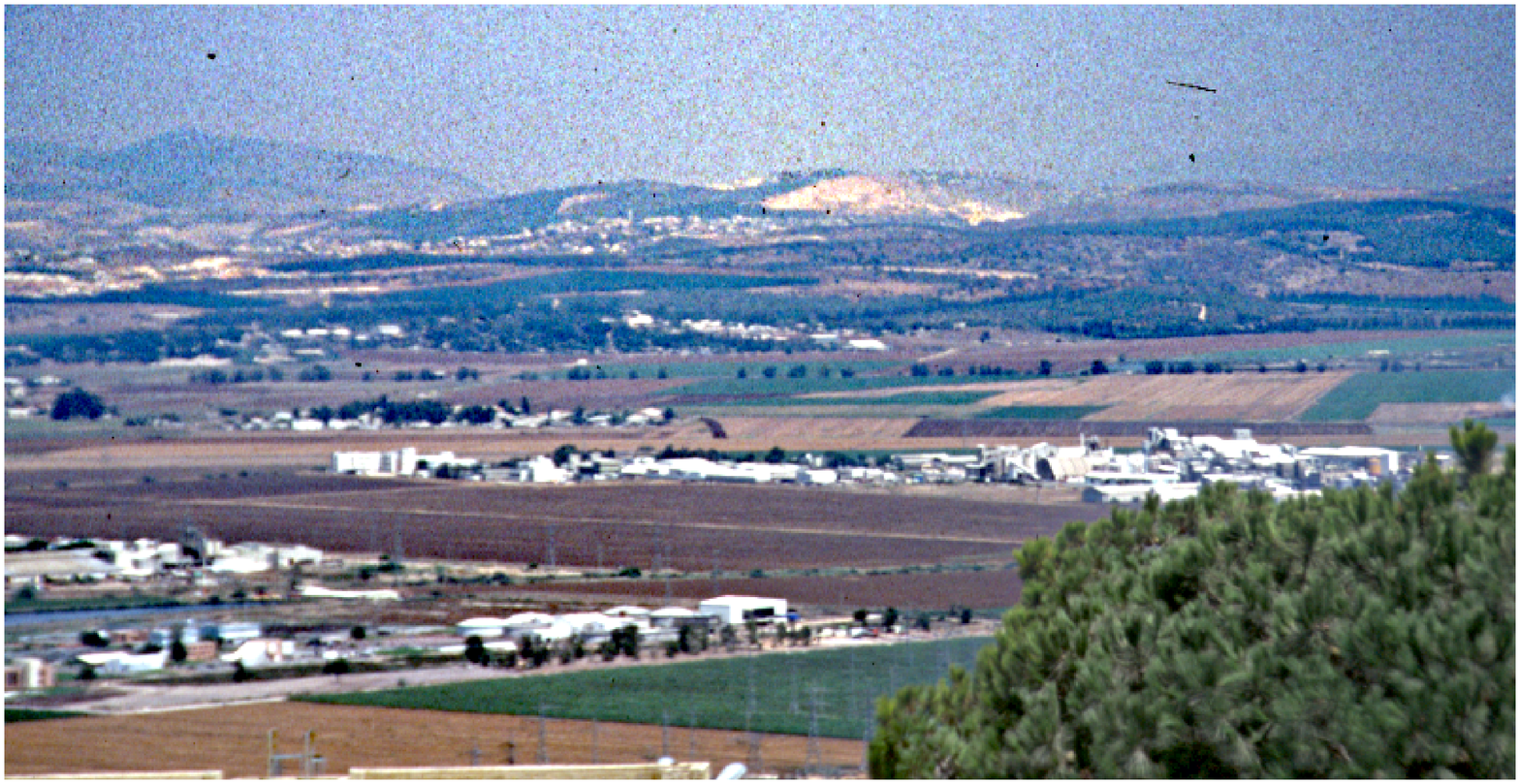} \\
  \small (a) Input & \small (b) Result of \cite{HeST09} \\
  \includegraphics[width=0.485\linewidth]{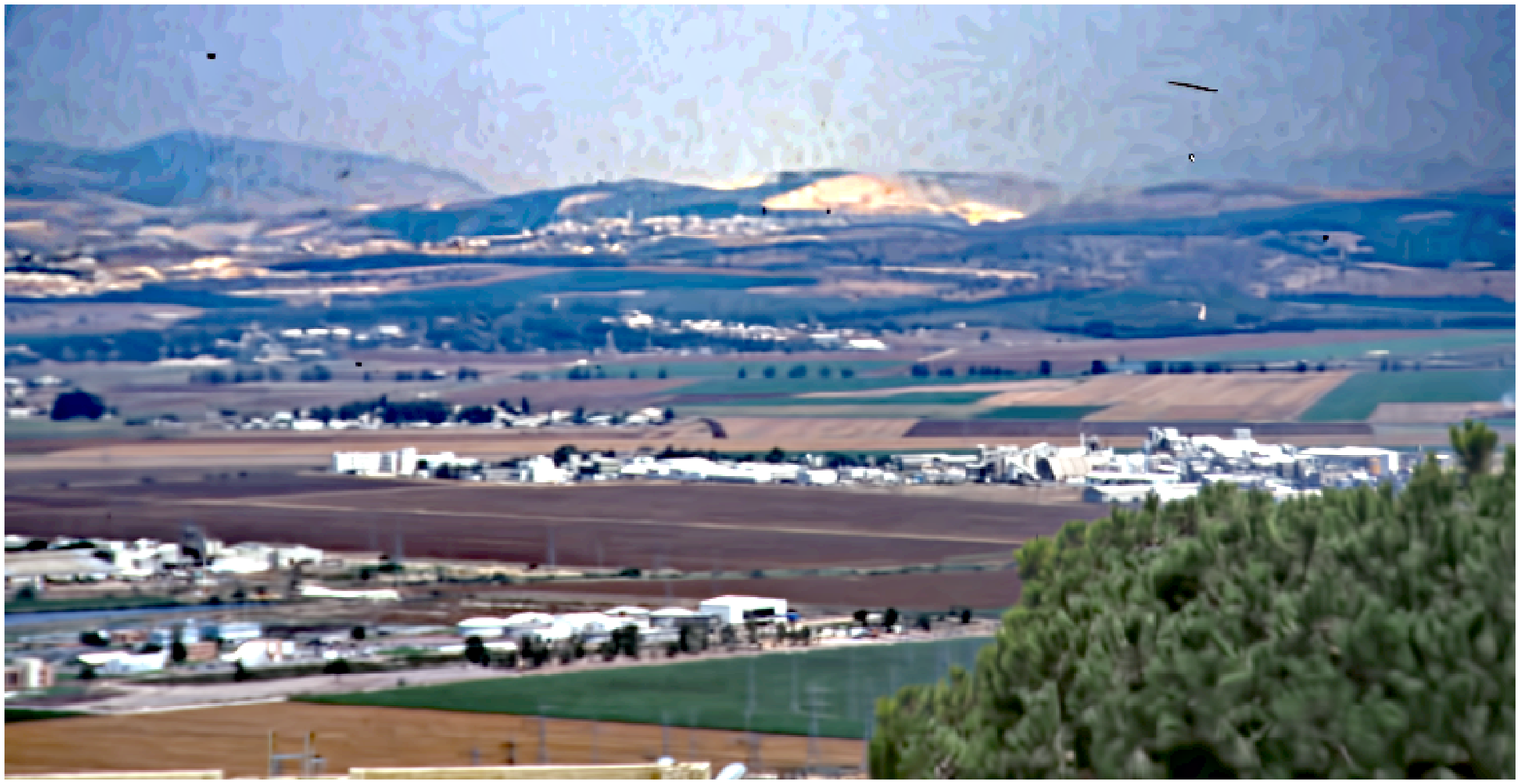} &
  \includegraphics[width=0.485\linewidth]{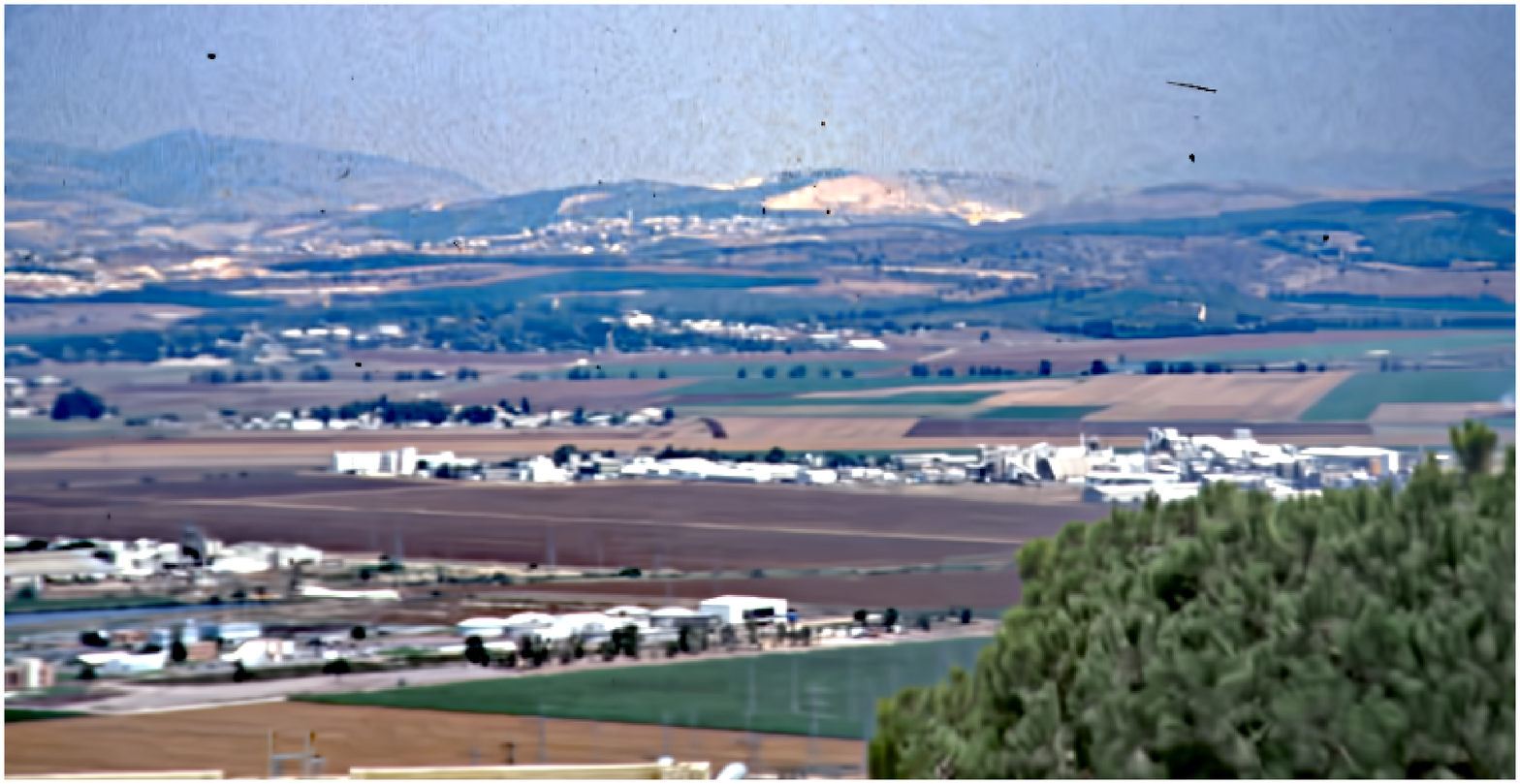} \\
  \small (c) Denoise (BM3D with $\sigma=5$) before dehazing \cite{HeST09}  & \small (d) Result by denoising (b) (BM3D with $\sigma=15$) \\
  \includegraphics[width=0.485\linewidth]{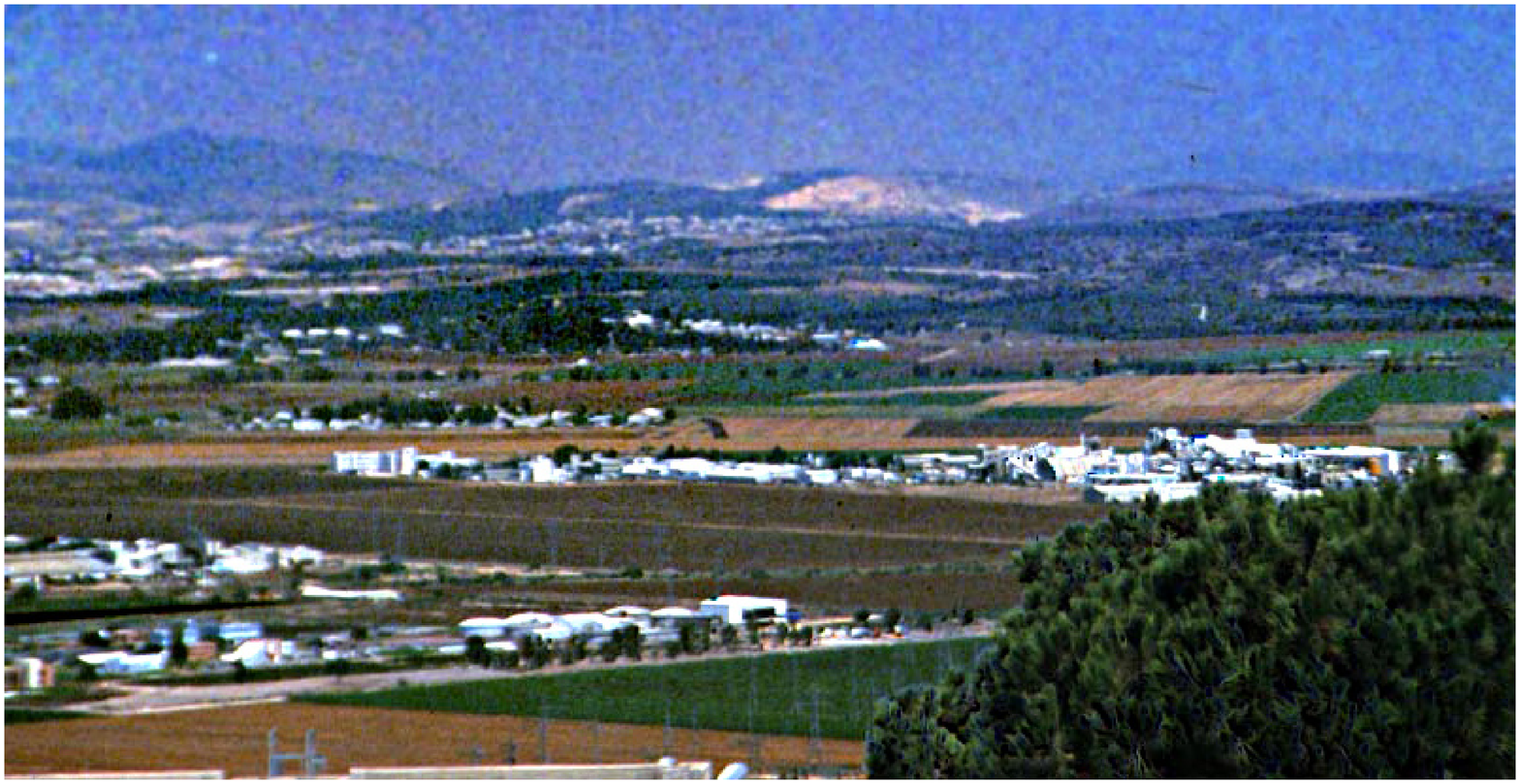} &
  \includegraphics[width=0.485\linewidth]{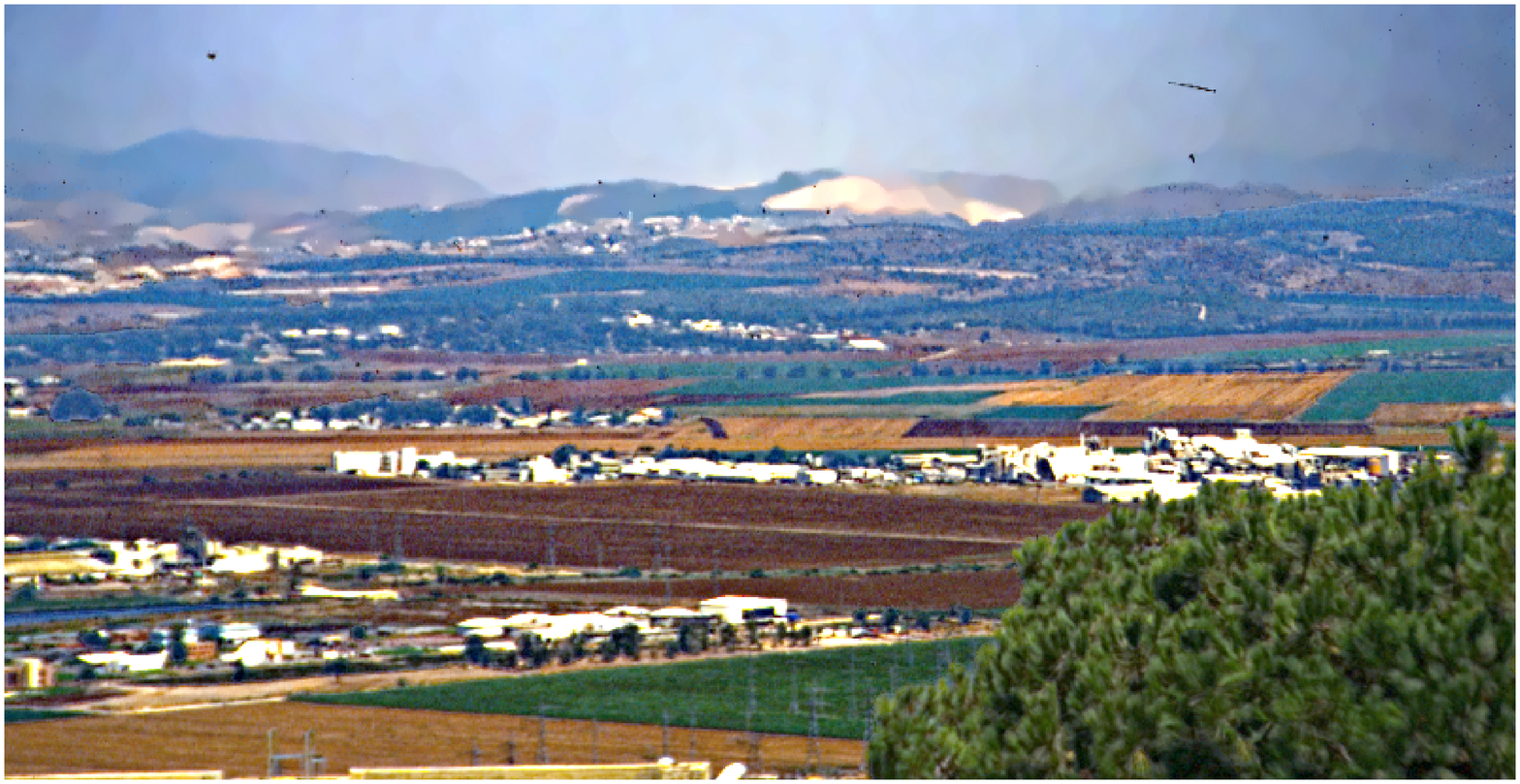} \\
  \small (e) Result of \cite{SchechnerA07} & \small (f) Our result \\
\end{tabular}
\begin{tabular}{@{\hspace{0.0mm}}c@{\hspace{1mm}}c@{\hspace{1mm}}c@{\hspace{1mm}}c@{\hspace{1mm}}c@{\hspace{1mm}}c@{\hspace{0mm}}}
  \includegraphics[width=0.158\linewidth]{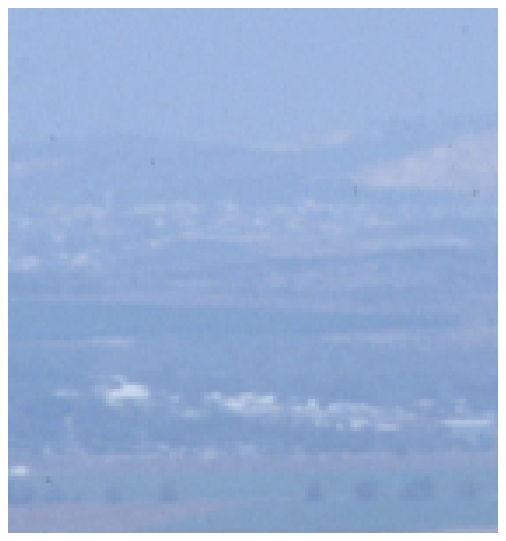} &
  \includegraphics[width=0.158\linewidth]{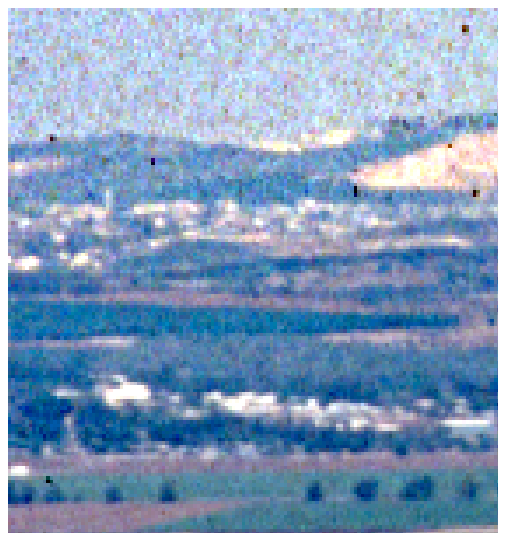} &
  \includegraphics[width=0.158\linewidth]{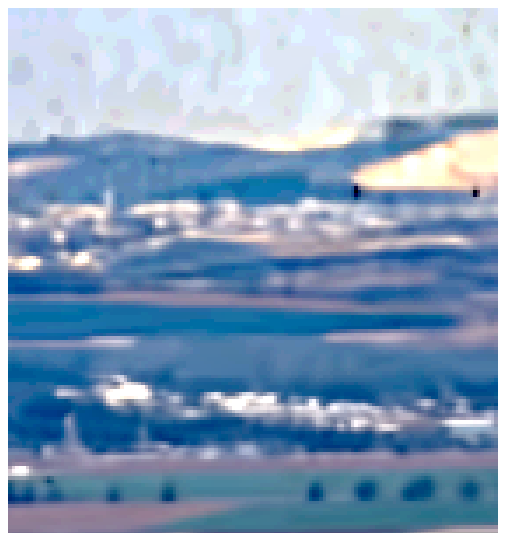} &
  \includegraphics[width=0.158\linewidth]{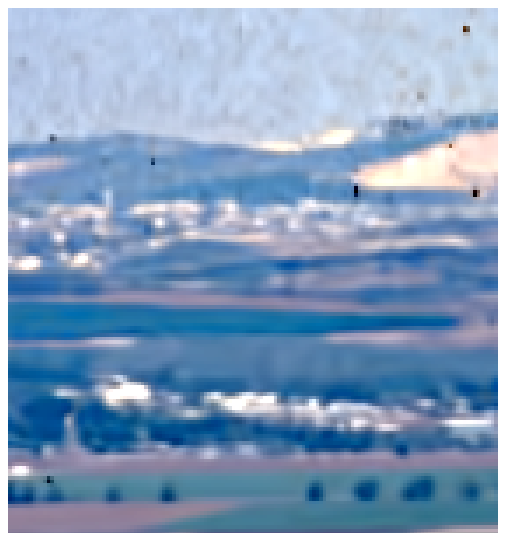} &
  \includegraphics[width=0.158\linewidth]{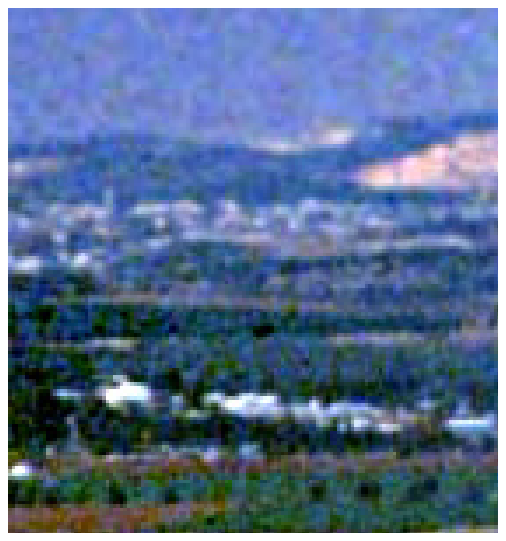} &
  \includegraphics[width=0.158\linewidth]{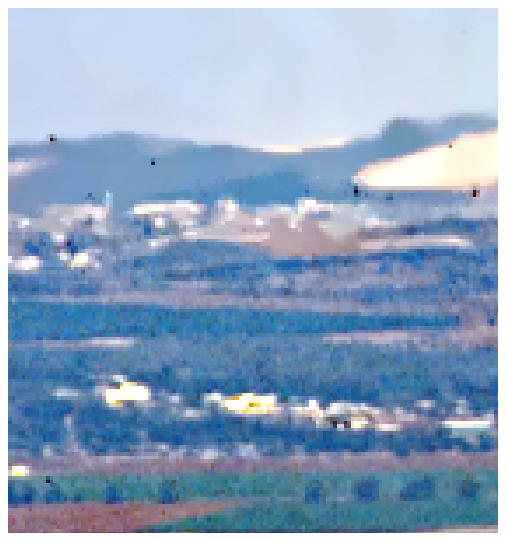} \\
  \small (g) & \small (h) & \small (i) & \small (j) & \small (k) & \small (l)
\end{tabular}
\caption{ A dehazing example. Given input (a), our method produces the result (f). It not only removes intensive noise, also retains a great amount of details as well. Close-ups are shown in the last row.} \label{fig:denoise_cmp_large}
\end{figure*}

\subsection{Comparison on transmission map}
Our non-local total variation smoothness term, working in concert with our point-wise data fidelity, is able to preserve thin structures comparing to the patch-based prior define in \cite{HeST09}. One example is shown in Fig. \ref{fig:pano2}. (e) is the transmission map of \cite{HeST09}, where inaccurate structure boundaries exist. They induce halos to the result shown in (c). Our method, shown in (f), preserves the structural edges. The corresponding restored image is in (d). Close-ups are shown in (b). We note that fine structures are very common in natural scenes. Their restoration is thus vital in scattering media removal.

\begin{figure*}[bpth]
\begin{tabular}{@{\hspace{0mm}}c@{\hspace{1mm}}c@{\hspace{0mm}}}
  \includegraphics[width=0.495\linewidth]{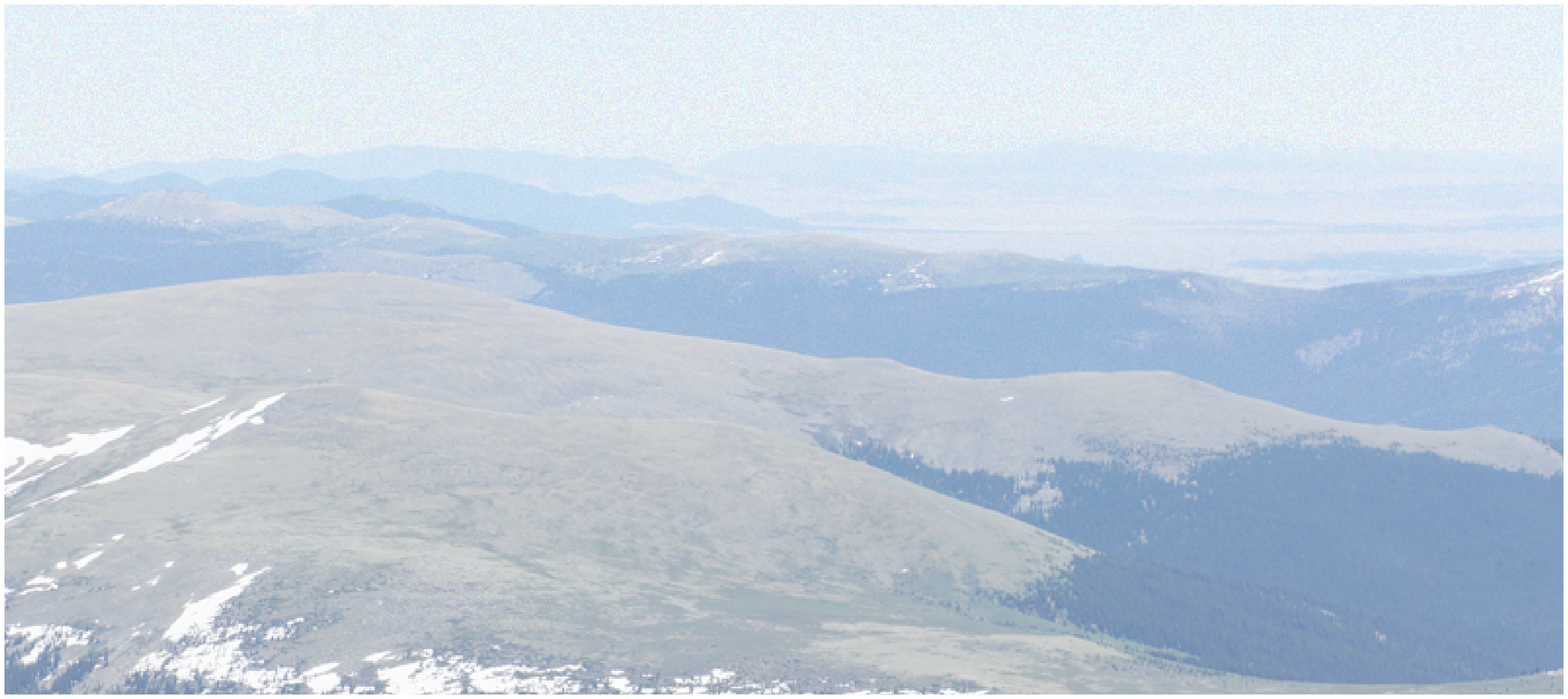} &
  \includegraphics[width=0.495\linewidth]{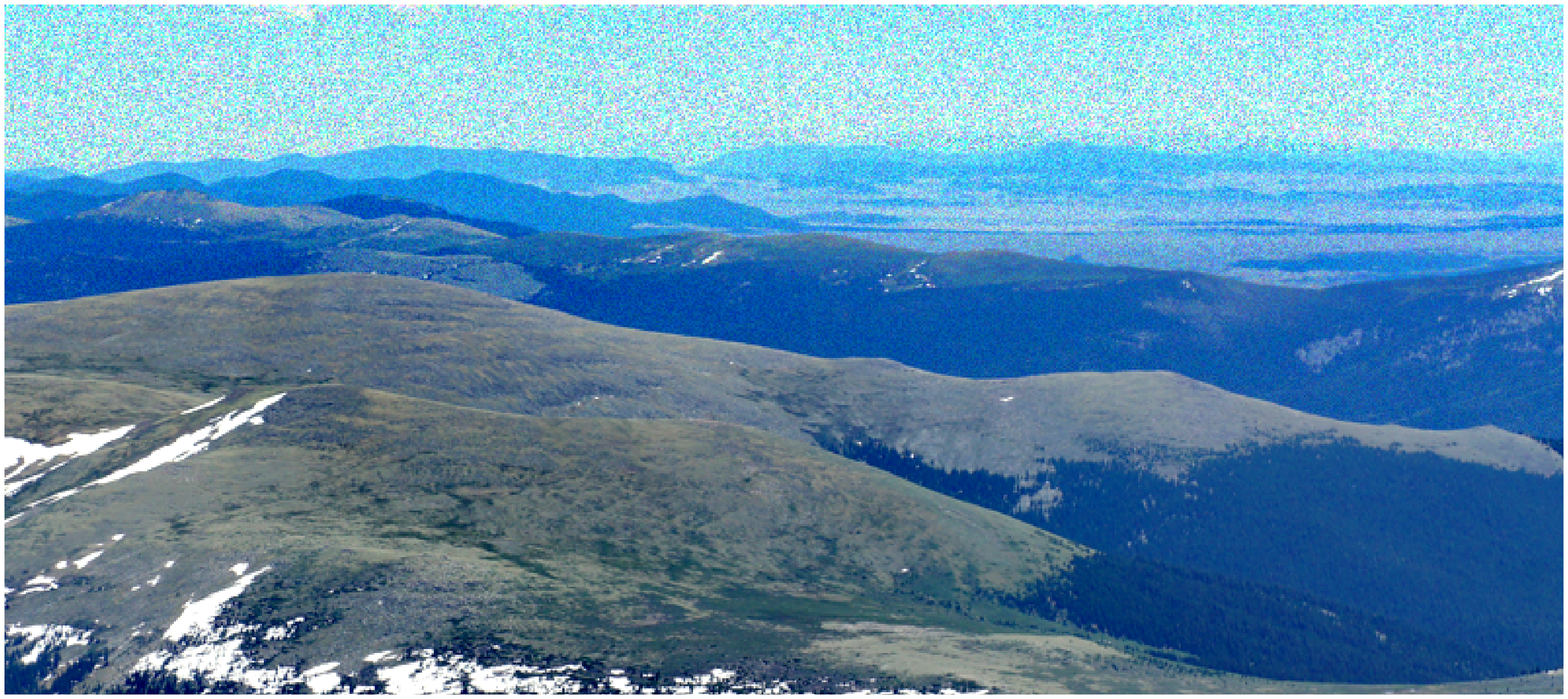}  \\
  { (a) Input} &
  { (b) Result of \cite{HeST09}} \\
  \includegraphics[width=0.495\linewidth]{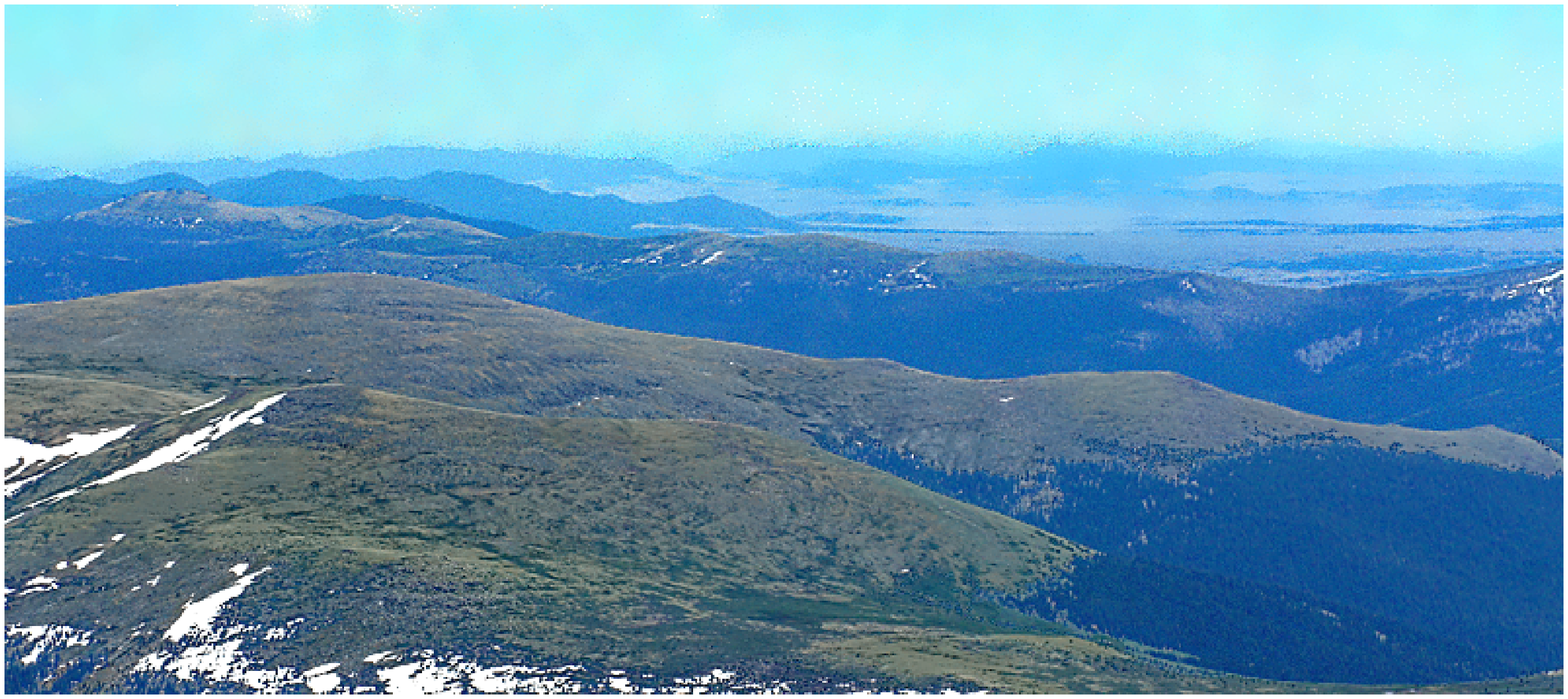} &
  \includegraphics[width=0.495\linewidth]{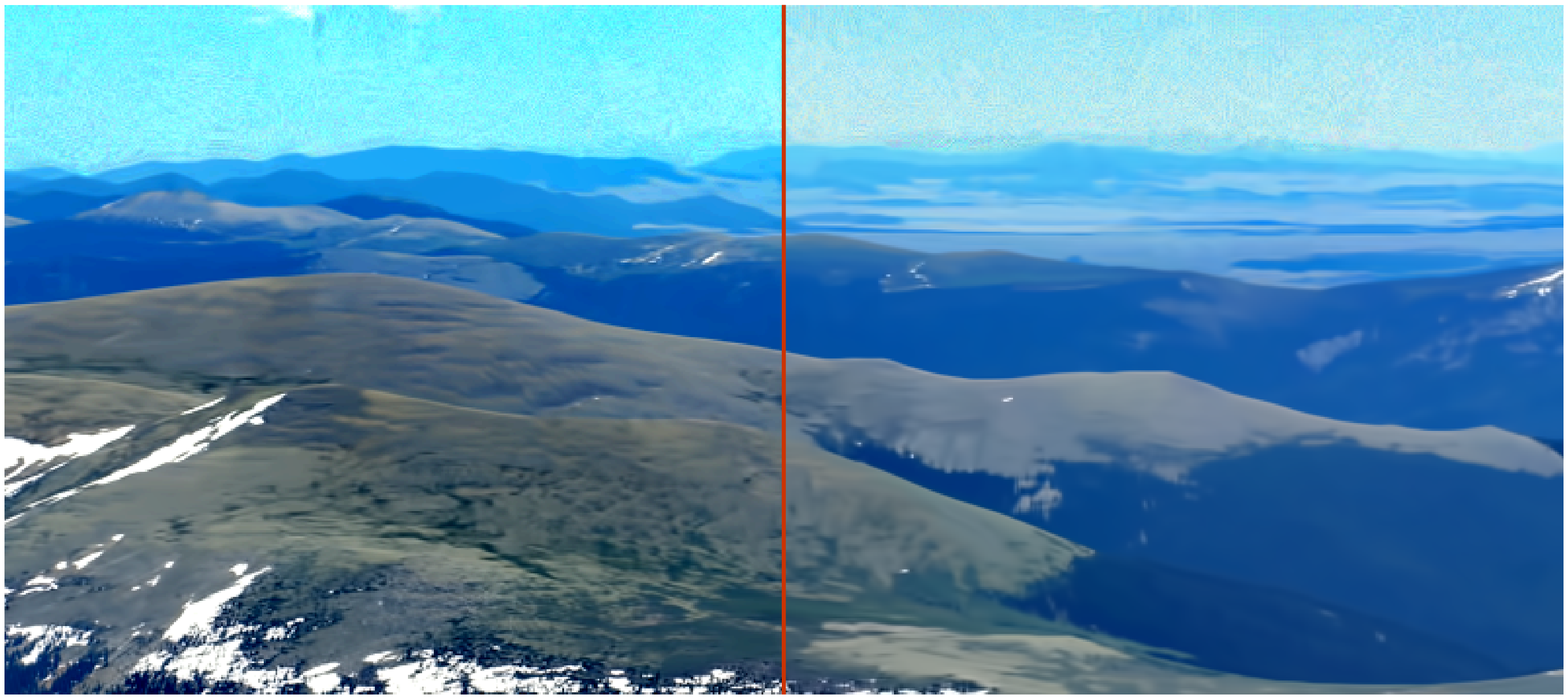}  \\
  { (c) Our result} &
  { (d) Denoising performed before (left) } \\
  & { and after (right) dehazing}
\end{tabular}\vspace{0.05in}
\caption{ Heavy fog lifting. For the challenging input image (a), the state-of-the-art method produces the result shown in (b). Intuitively applying denoising before and after dehazing also cannot produce reasonable results, as shown in (d). Our result in (c) contains most structural details with noise being suppressed.}
\label{fig:denoise_mount}
\end{figure*}

\begin{figure*}[tb]
\begin{minipage}{0.49\linewidth}
\centering
\begin{tabular}{@{\hspace{0mm}}c@{\hspace{1mm}}c@{\hspace{0mm}}}
  \includegraphics[width=0.49\linewidth]{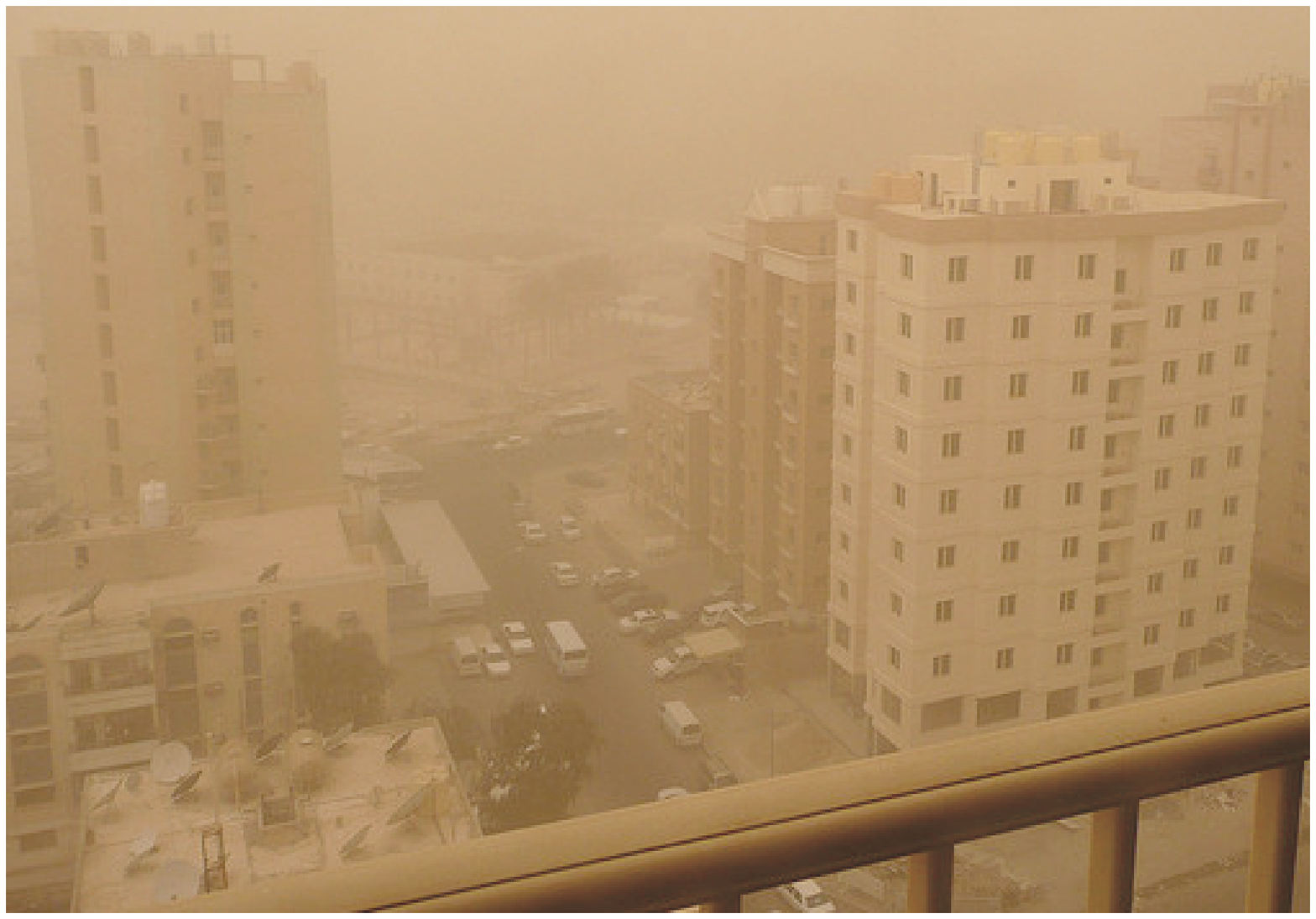}&
  \includegraphics[width=0.49\linewidth]{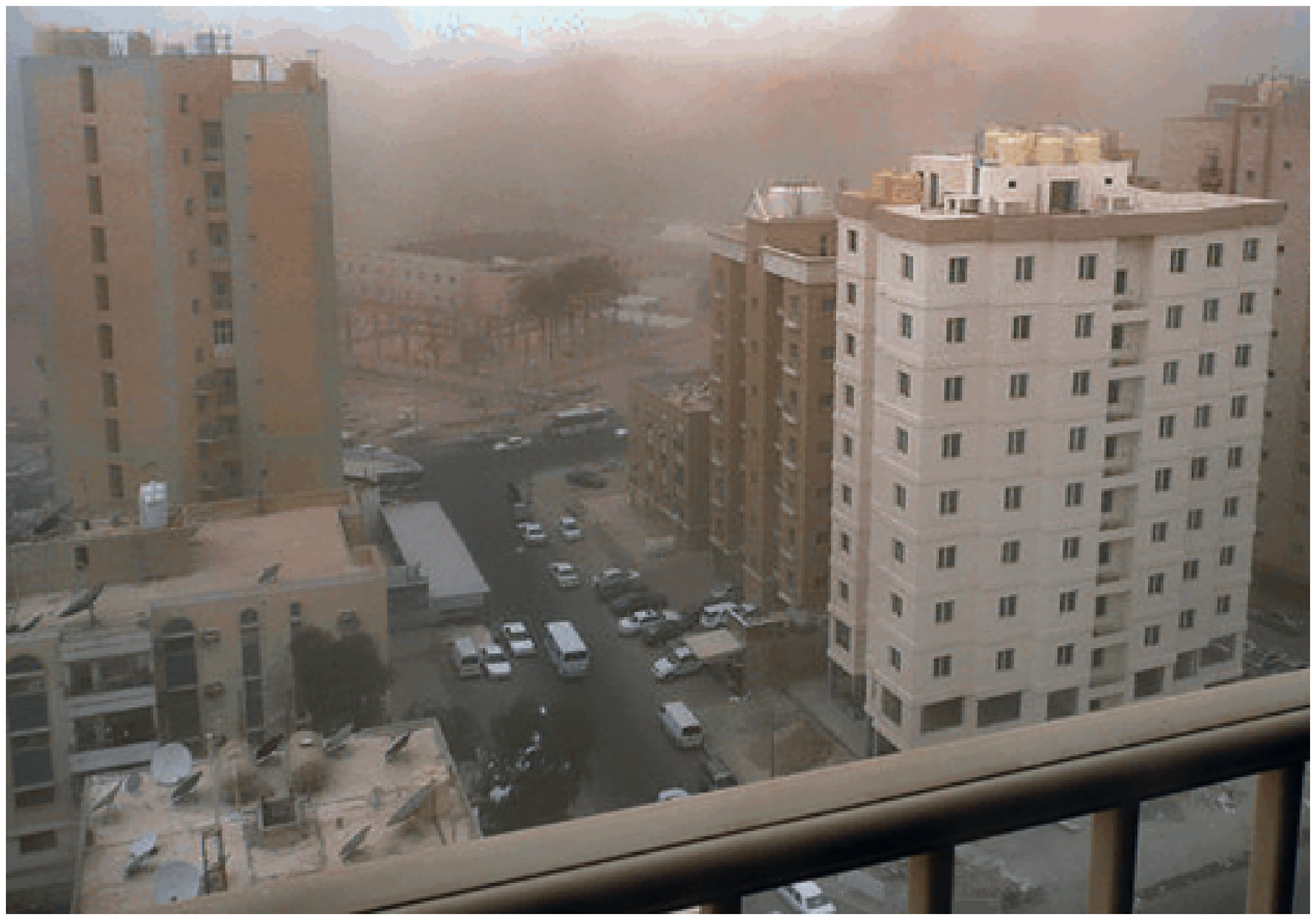}   \\
 \includegraphics*[viewport=20 20 400 280,width = 0.49\linewidth]{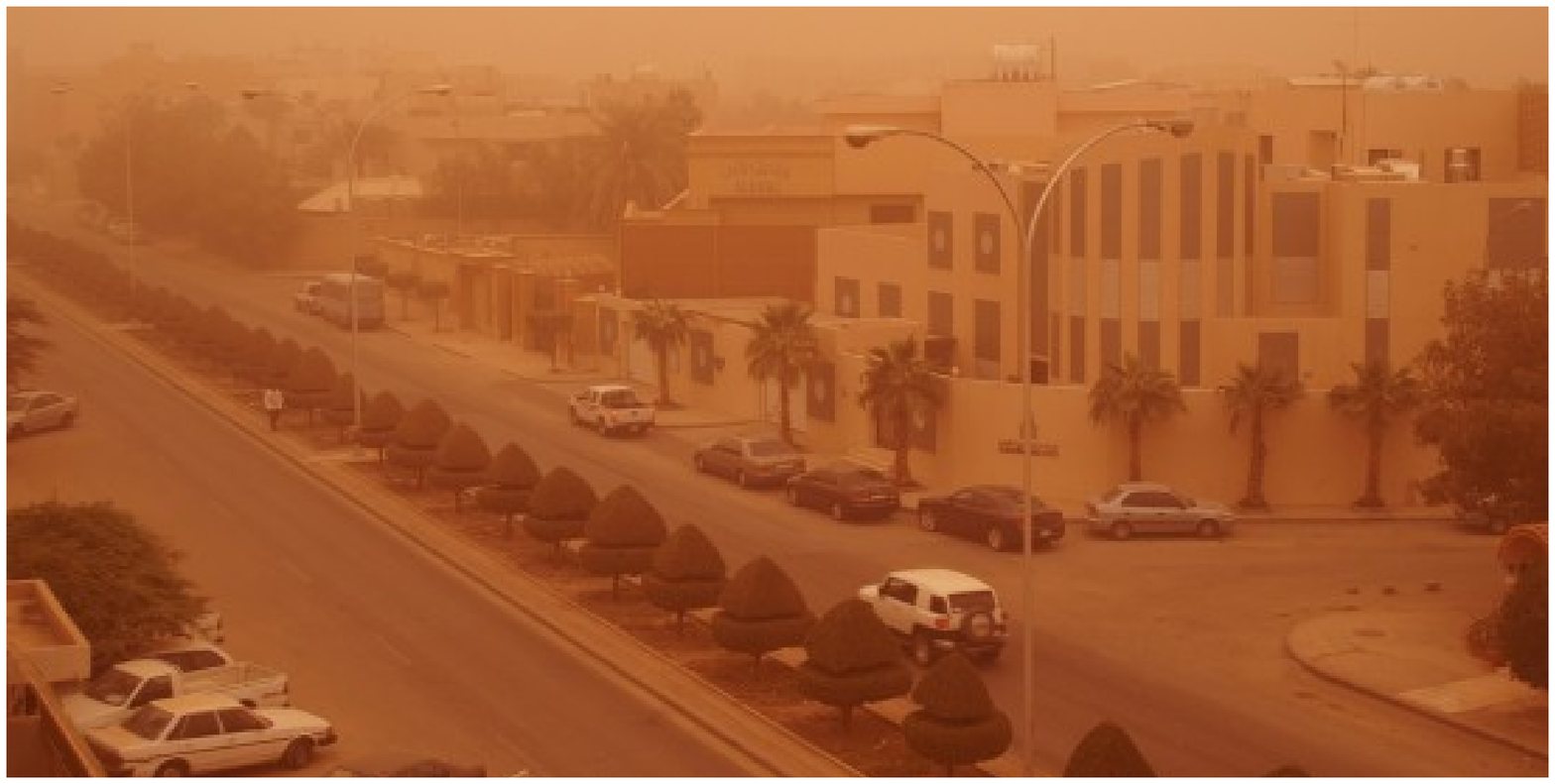} &
 \includegraphics*[viewport=20 20 400 280,width = 0.49\linewidth]{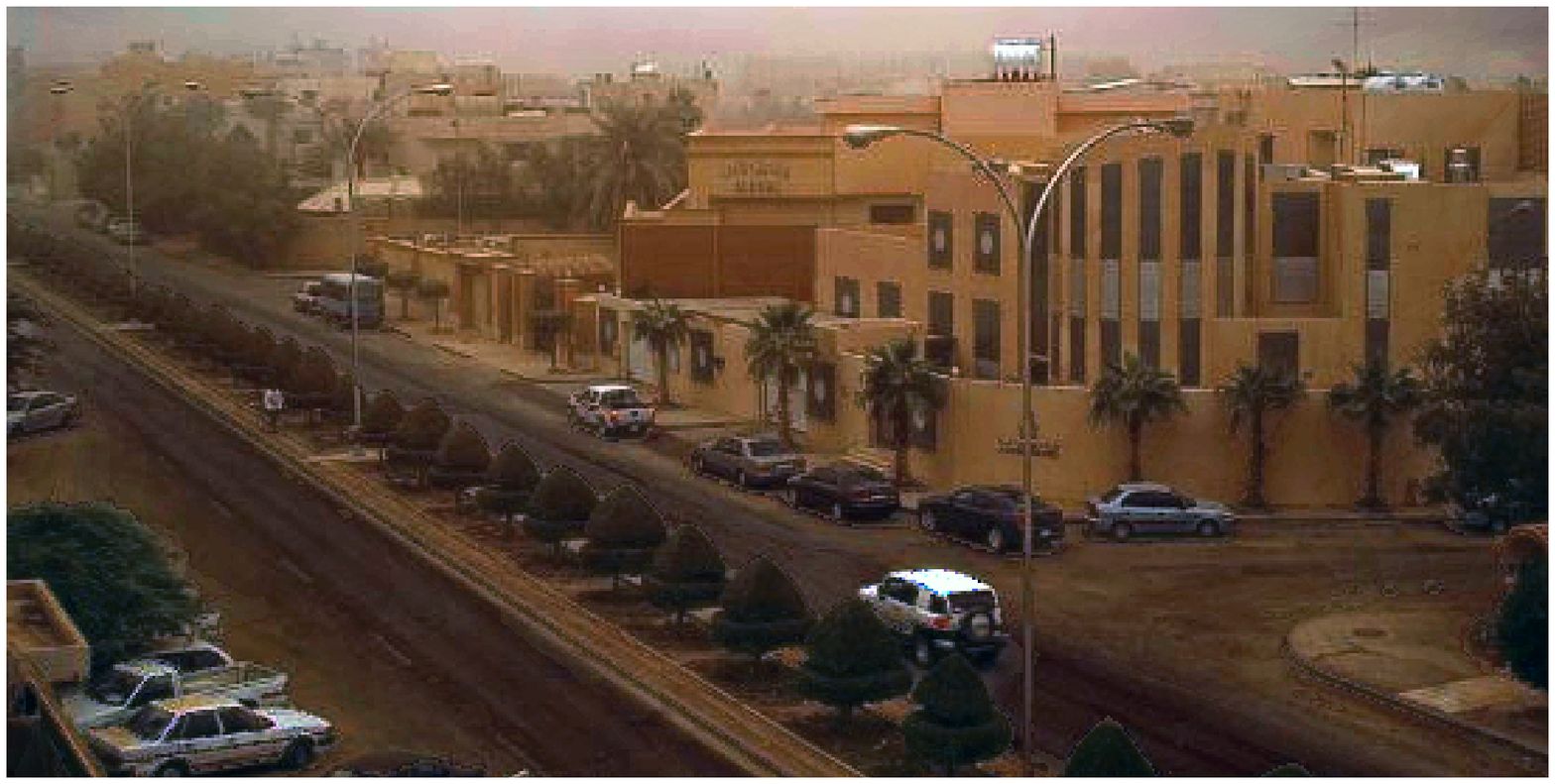} \\
  (a) Input & (b)  Our result
\end{tabular}
\end{minipage}
\begin{minipage}{0.005\linewidth}
\centering
~
\end{minipage}
\begin{minipage}{0.5\linewidth}
\centering
\begin{tabular}{@{\hspace{0mm}}c@{\hspace{1mm}}c@{\hspace{0mm}}}
 \includegraphics*[viewport=0 20 208 165,width=0.49\linewidth]{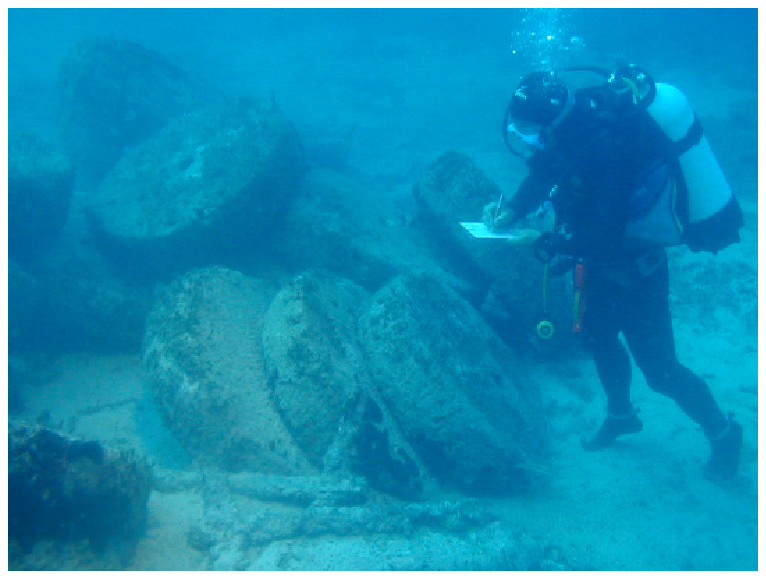} &
 \includegraphics*[viewport=0 20 208 165,width=0.49\linewidth]{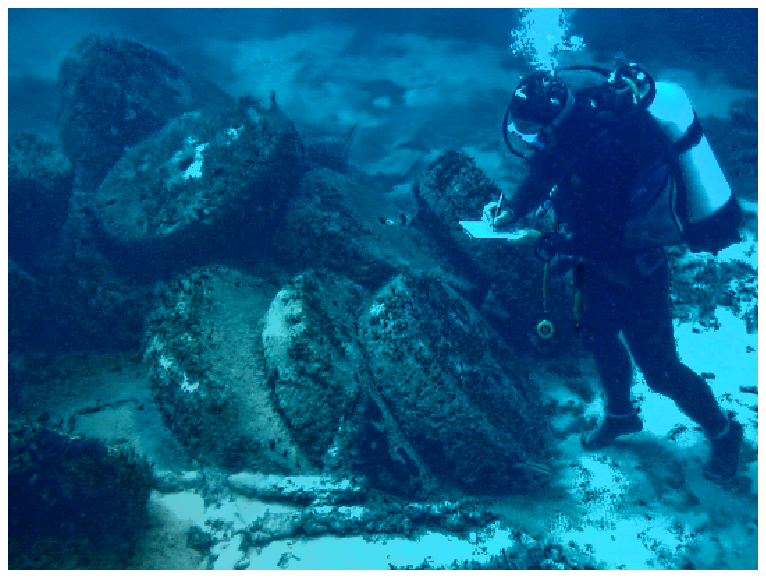} \\
 \includegraphics[width = 0.49\linewidth]{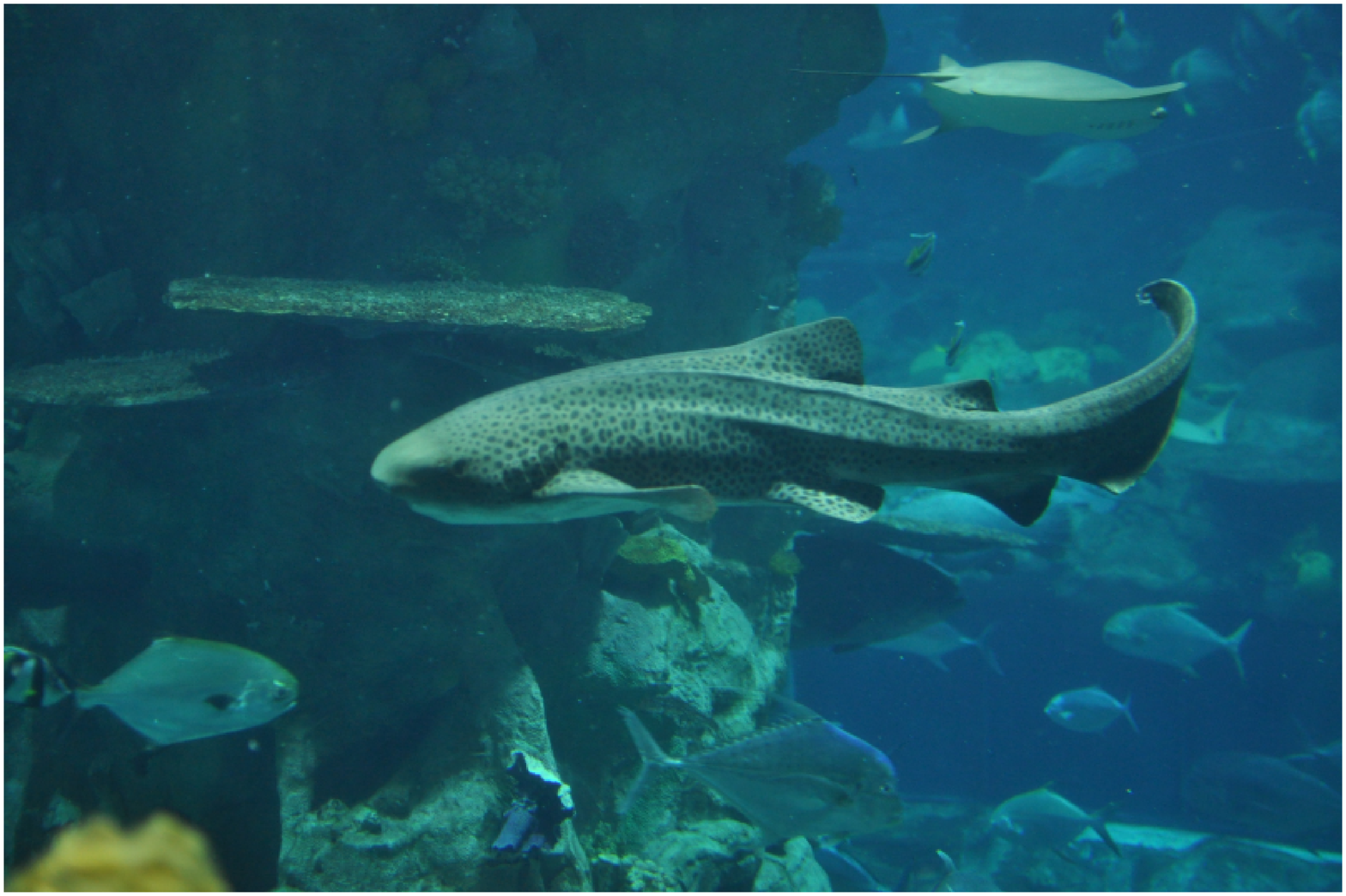} &
 \includegraphics[width = 0.49\linewidth]{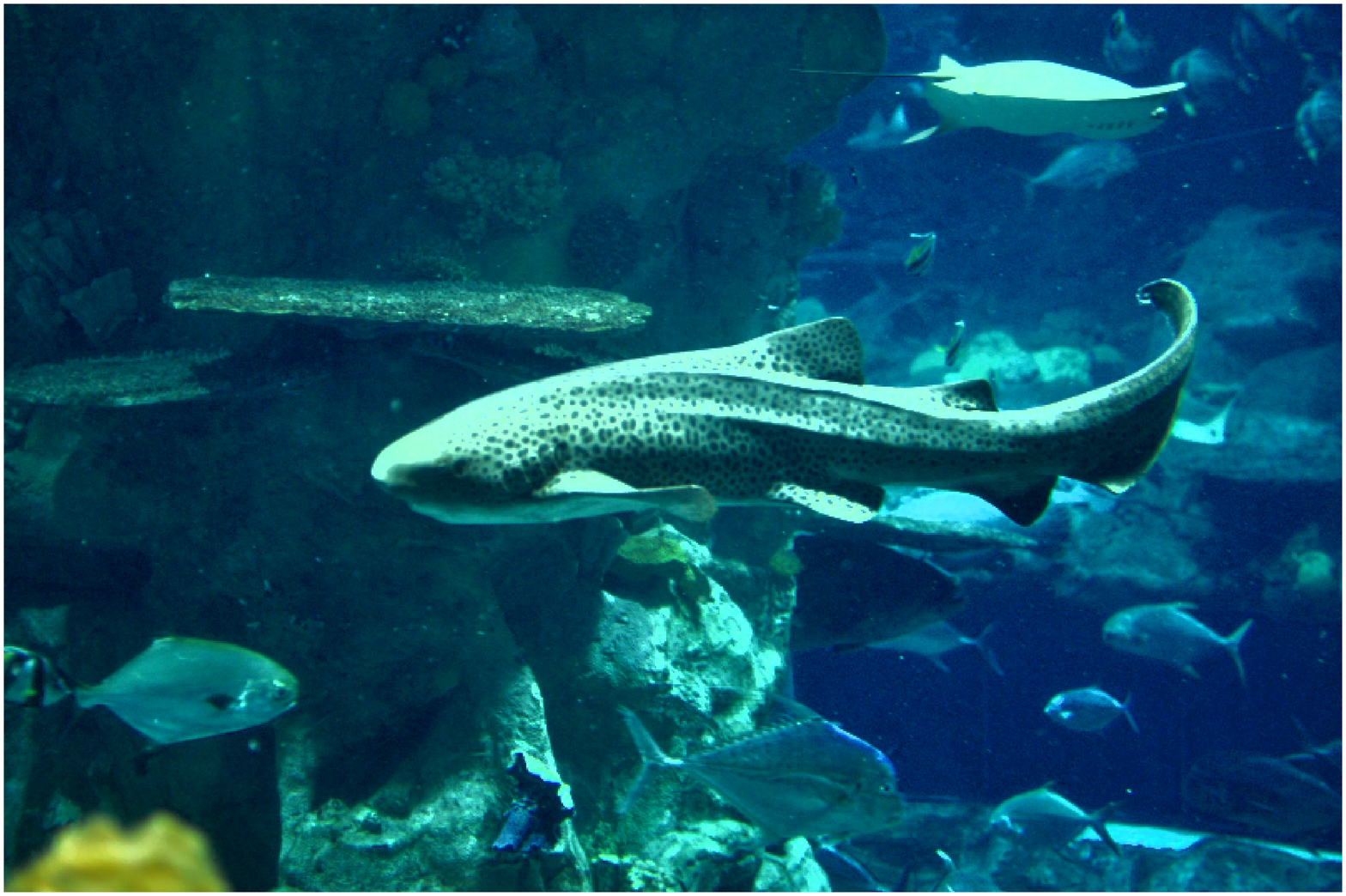} \\
  (c) Input & (d)  Our result
\end{tabular}
\end{minipage}\vspace{0.05in}
\caption{ {Underwater and dust storm picture restoration.}}\label{fig:sand}
\end{figure*}

\subsection{Comparison on noise reduction}
We exhibit our advantageous ability in handling significant noise. We compare our result with a previous regularized restoration approach \cite{SchechnerA07}, as well as strategies applying denoising before and after layer removal. One example is shown in Fig. \ref{fig:denoise_cmp_large}. In the result of \cite{HeST09} (b) unnoticeable noise in the original image is greatly enhanced.  The result shown in (c), which performs denoising before dehazing, still contains much noise even with spatial regularization. Denoising after dehazing, on the other hand, can hardly remove intensive noise out of the latent image structure. As shown in (d), even state-of-the-art BM3D denoising method destroys many latent image details, while considerable noise in the sky is left over. (e) is the result of \cite{SchechnerA07} with a local TV regularizer. (f) is our final result. Noise is suppressed while underlaying structures are well preserved. Close-ups in the last row are from (a)-(f) respectively. Another example on noise reduction is shown in Fig. \ref{fig:denoise_mount}, demonstrating the effectiveness of our transmission-aware regularization.

\subsection{General scattering medium removal}
Our restoration method makes no assumption on the scattering media, making it applicable to fog, dust storm, and underwater pictures. The example in Fig. \ref{fig:sand}(a) consists of a dense sand layer and is restored by our method in (b). For underwater environment, however, applying our method to all image channels results inaccurate color, since the red channel is generally much weaker than the other two due to the absorbtion by water. We choose to apply our method to the intensity channel only. An example is shown in Fig. \ref{fig:sand}(c)-(d). A large amount of structural information is recovered and contrast is greatly enhanced.

\section{Concluding Remarks}
We have presented a new model for scattering media layer removal from a single image. We introduced the transmission lower-bound condition and provided a very effective optimization framework incorporating several novel terms to solve the challenging noise amplification and depth estimation problems. Our method applied to images taken in fog, sandstorm, and underwater scenes.

\appendix

\section{Numerical Solver}
In this section we describe our the numerical solvers (referring to Sections \ref{sec:transmission} and \ref{sec:latent_image} in the paper) to compute the depth $D$ and the latent clear image $L$ respectively.

\subsection{Depth $D$ Solver} \label{sec:est_D}
Depth $D$ is computed by minimizing Eq. (\ref{eq:energy_d}) in the paper, where $E(D)$ is written as
\begin{align}\label{eq:energy_D}
E(D) = & \sum_{{\mathrm{x}}}\sum_c|D(\mathrm{x})-(\bar{i}^c(\mathrm{x})-\bar{l}^c(\mathrm{x}))|^2 + \nonumber \\
&~~ \lambda \sum_{\mathrm{x}}\sum_{\mathrm{y}\in W(\mathrm{x})} \tilde{w}_d(\mathrm{x},\mathrm{y})|D({\mathrm{x}})-D({\mathrm{y}})|,
\end{align}
where $\bar{i}^c(\mathrm{x})$ and $\bar{l}^c(\mathrm{x})$ are the observed and latent images in the logarithm domain. $\tilde{w}_d(\mathrm{x},\mathrm{y})$ is a structure-aware penalty function defined in the paper. Directly minimizing Eq. (\ref{eq:energy_D}) is difficult as the non-local smoothness term \cite{PeyreBC08} involves a number of neighboring pixels, making gradient descent work poorly. We instead employ an iterative relaxation scheme to compute $D$.

Initially, we set $D^{(0)} = v$ as the start point. In the $\tau$-th iteration, we minimize the energy function
\begin{align}\label{eq:energy_Dt}
E(D^{(\tau)}) =  &
\sum_{{\mathrm{x}}}\sum_c|D^{(\tau)}(\mathrm{x})-(\bar{i}^c(\mathrm{x})-\bar{l}^c(\mathrm{x}))|^2
+ \nonumber\\
&~~ \lambda \sum_{\mathrm{x}}\sum_{\mathrm{y}\in W(\mathrm{x})}
\tilde{w}_d(\mathrm{x},\mathrm{y})|D^{(\tau)}({\mathrm{x}})-D^{(\tau-1)}({\mathrm{y}})|,
\end{align}
where the superscript $(\tau)$ indexes iterations. In solving $D^{(\tau)}(\mathrm{x})$ for each $\mathrm{x}$, relaxation keeps other values fixed, which results in an efficient estimation process. By further denoting by $\{D^{(\tau-1)}_h\}_{h=0}^{W-1}$ the sorted $D^{(\tau-1)}$ within the window $W(\mathrm{x})$ in an ascending order and by $w_h$ the correspondingly sorted weight $\tilde{w}_d(\mathrm{x},\mathrm{y})$, $D^{(\tau)}(\mathrm{x})$ can be written as
\begin{align}\label{eq:median}
E(D^{(\tau)}(\mathrm{x})) = & \sum_c |D^{(\tau)}(\mathrm{x})-(\bar{i}^c(\mathrm{x})-\bar{l}^c(\mathrm{x}))|^2 + \nonumber \\
    &~~ \lambda \sum_{h=0}^{W-1} w_h | D^{(\tau)}(\mathrm{x}) - D^{(\tau-1)}_h |.
\end{align}
Eq. (\ref{eq:median}) has two essential features that enable efficient computation. One is that the data term is strictly convex and differentiable, with its derivative $2 \cdot \sum_c (D^{(\tau)}(\mathrm{x})-(\bar{i}^c(\mathrm{x})-\bar{l}^c(\mathrm{x})))$ a bijective function. The other property is that weight $w_h$ is non-negative for all $h$s by definition. With these properties and also according to the proof in \cite{Li_wmf09}, Eq. (\ref{eq:median}) can be efficiently optimized by computing the median value in
\begin{equation}
M = \left\{ D^{(\tau-1)}_0, \ldots, D^{(\tau-1)}_{W-1}, r_0, \ldots, r_W  \right\},
\end{equation}
where
\begin{equation}\label{eq:rhDef}
r_h = \frac {\sum_c (\bar{i}^c(\mathrm{x})-\bar{l}^c(\mathrm{x}))} {3} + \frac {\lambda} {6} \left( - \sum_{j=0}^{h-1} w_j + \sum_{j=h}^{W-1} w_j \right),
\end{equation}
for $h=0, \ldots, W$.

Based on this relaxation scheme, we present the overview of our algorithm in Algorithm \ref{tb:1}. Typically, $\lambda$ is set to {15}. Two or three passes, i.e., $n\leq 3$, to update all pixels,  are enough to yield good results in our experiments.

\begin{Algorithm}[t]{\linewidth}
\caption{Relaxation Procedure to Compute $D$.}\label{tb:1}
\begin{algorithmic}[1]
\STATE {\bf input:} initial depth $D^{(0)}=v$; the maximum number of iterations $n$
\STATE $\tau = 1$.
\STATE Apply median filter to $D^{(\tau-1)}$ according to Eq. (\ref{eq:median}), and get $D^{(\tau)}$.
\STATE $\tau = \tau+1$. If $\tau \leq n$, go to Step 2.
\STATE {\bf output:} $D$
\end{algorithmic}
\end{Algorithm}

\begin{Algorithm}[t]{\linewidth}
\caption{Relaxation Procedure to Compute $L$.}\label{tb:2}
\begin{algorithmic}[1]
\STATE {\bf input:} initial image $L^{(0)}=L_0$; the maximum number of iteration $n$
\STATE $\tau = 1$.
\STATE Apply median filter to $L^{(\tau-1)}$ according to Eq. (\ref{eq:median2}), which yields $L^{(\tau)}$.
\STATE $\tau = \tau+1$. If $\tau \leq n$, go to Step 2.
\STATE {\bf output:} $L$
\end{algorithmic}
\end{Algorithm}

\subsection{Optimizing Latent Image $L$}\label{sec:est_L}
The energy function for solving $L$ (in Section 2.2 in the paper) is similar to Eq. (\ref{eq:energy_D}). We decompose it into channels $c\in\{r,g,b\}$, and write it as
\begin{align}\label{eq:energy_L}
E(L^c) = & \sum_{{\mathrm{x}}} t(\textrm{x})^2 |L^c(\textrm{x}) - L_0^c(\textrm{x})|^2 + \nonumber\\
&~~ \lambda_L \sum_{\mathrm{x}}\sum_{\mathrm{y}\in W(\mathrm{x})} \bar{m}(\mathrm{x},\mathrm{y})|L^c({\mathrm{x}})-L^c({\mathrm{y}})|.
\end{align}
By applying similar relaxation, we get the following pixel-wise energy minimization problem
\begin{align}\label{eq:median2}
E({L^c}^{(\tau)}(\mathrm{x})) = & ~ t(\textrm{x})^2|{L^c}^{(\tau)}(\mathrm{x})-L_0^c(\textrm{x})|^2 + \nonumber\\
&~~ \lambda_L \sum_{h=0}^{W-1} m_h | {L^c}^{(\tau)}(\mathrm{x}) - {L^c_h}^{(\tau-1)} |.
\end{align}
It can also be solved by finding the median value in
\begin{equation}
\left\{ {L^c_0}^{(\tau-1)}, \ldots, {L^c_{W-1}}^{(\tau-1)}, z_0, \ldots, z_W \right\},
\end{equation}
where
\begin{equation}
z_h = L_0^c(\textrm{x}) + \frac {\lambda_L} {2 t(\textrm{x})^2} \left( - \sum_{j=0}^{h-1} m_j + \sum_{j=h}^{W-1} m_j \right),
\end{equation}
for $h=0, \ldots, W$.

Our algorithm is shown in Algorithm \ref{tb:2}. $\lambda_L$ usually varies from 0.001 to 0.02 to prevent over-smoothing. Iteration number $n=1\sim 3$ is enough.

\bibliographystyle{ieee}
\bibliography{dehaze}

\end{document}